\documentclass[journal]{IEEEtran}
\usepackage{amsmath,amsfonts}
\usepackage{algorithmic}
\usepackage{algorithm}
\usepackage{array}
\usepackage[caption=false,font=normalsize,labelfont=sf,textfont=sf]{subfig}
\usepackage{textcomp}
\usepackage{stfloats}
\usepackage{url}
\usepackage{verbatim}
\usepackage{graphicx}
\usepackage{cite}
\usepackage{balance}
\usepackage{tabularx}
\usepackage{array}
\usepackage[tracking=true]{microtype}
\newcolumntype{C}[1]{>{\centering\arraybackslash}p{#1}}
\usepackage{float}
\usepackage{booktabs}
\usepackage{multirow}

\newcolumntype{Y}{>{\raggedright\arraybackslash}X} %

\begin{document}

\title{Communication-Free Collective Navigation for a Swarm of UAVs \\ via LiDAR-Based Deep Reinforcement Learning}

\author{Myong-Yol Choi, Hankyoul Ko, Hanse Cho, Changseung Kim, Seunghwan Kim, Jaemin Seo, \\ and Hyondong Oh

\thanks{This research was supported by Theater Defense Research Center funded by Defense Acquisition Program Administration under Grant UD200043CD, National Research Foundation of Korea (NRF) grant funded by the Korea government (MSIT) (2023R1A2C2003130), and Unmanned Vehicles Core Technology Research and Development Program through the National Research Foundation of Korea (NRF) and Unmanned Vehicle Advanced Research Center (UVARC) funded by the Ministry of Science and ICT, the Republic of Korea (2020M3C1C1A01082375). \textit{(Myong-Yol Choi led the project, and Myong-Yol Choi and Hankyoul Ko contributed equally to this work.)(Corresponding author: Hyondong Oh.)}}

\thanks{\textls[-30]{M.-Y. Choi, H. Ko, H. Cho, C. Kim, S. Kim, and J. Seo are with the Department of Mechanical Engineering, Ulsan National Institute of Science and Technology, Ulsan, Korea (e-mail: mychoi@unist.ac.kr; kyoul@unist.ac.kr; joahdzl@unist.ac.kr; pon02124@unist.ac.kr; kevin6960@unist.ac.kr; qkek1019@unist.ac.kr).}}
\thanks{H. Oh is with the Department of Mechanical Engineering, Korea Advanced Institute of Science and Technology, Daejeon, Korea (e-mail: h.oh@kaist.ac.kr).}}


\maketitle


\begin{abstract}
This paper presents a deep reinforcement learning (DRL) based controller for collective navigation of unmanned aerial vehicle (UAV) swarms in communication-denied environments, enabling robust operation in complex, obstacle-rich environments. Inspired by biological swarms where informed individuals guide groups without explicit communication, we employ an implicit leader-follower framework. In this paradigm, only the leader possesses goal information, while follower UAVs learn robust policies using only onboard LiDAR sensing, without requiring any inter-agent communication or leader identification. Our system utilizes LiDAR point clustering and an extended Kalman filter for stable neighbor tracking, providing reliable perception independent of external positioning systems. The core of our approach is a DRL controller, trained in GPU-accelerated Nvidia Isaac Sim, that enables followers to learn complex emergent behaviors—balancing flocking and obstacle avoidance—using only local perception. This allows the swarm to implicitly follow the leader while robustly addressing perceptual challenges such as occlusion and limited field-of-view. The robustness and sim-to-real transfer of our approach are confirmed through extensive simulations and challenging real-world experiments with a swarm of five UAVs, which successfully demonstrated collective navigation across diverse indoor and outdoor environments without any communication or external localization.
\end{abstract}


\section{Introduction}

\subsection{Background and Motivation}

\IEEEPARstart{U}{nmanned} aerial vehicle (UAV) swarms have demonstrated remarkable potential across diverse applications including search and rescue, surveillance, environmental monitoring, and precision agriculture due to their mobility, flexibility, and ability to access hard-to-reach areas~\cite{floreano2015science}. By leveraging the complementary sensing capabilities and spatial coverage of multiple agents, UAV swarms can accomplish complex missions that are difficult or impossible for individual UAVs to achieve efficiently~\cite{alqudsi2025uav}. Collective navigation, where multiple UAVs move together toward common destinations while maintaining group cohesion, has emerged as a key enabling capability for these applications~\cite{dorigo2020reflections}.

The foundations of collective motion have been extensively studied across multiple disciplines. Reynolds~\cite{reynolds1987flocks} pioneered distributed behavioral models using three fundamental rules: collision avoidance, velocity matching, and flock centering. Vicsek and Zafeiris~\cite{vicsek2012collective} later demonstrated that large-scale order can emerge from a much simpler rule: local velocity matching with noise. Olfati-Saber~\cite{olfati2006flocking} provided rigorous mathematical analysis introducing collective potential functions and $\alpha$-lattice structures for multi-agent coordination.

\begin{figure}[t]
    \centering
    \includegraphics[width=\columnwidth]{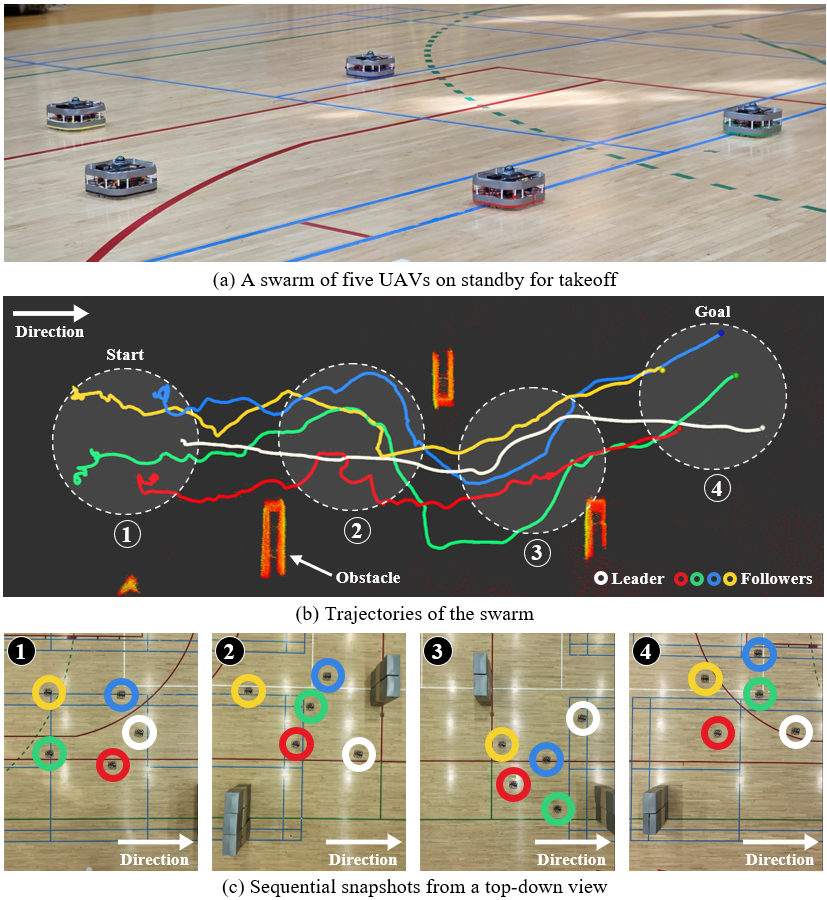}
    \vspace{-20pt}
    \caption{An example of real-world validation of the proposed LiDAR-based collective navigation. (a) A swarm of five UAVs on standby for takeoff. (b) Trajectories of the swarm reconstructed from onboard LiDAR. (c) Sequential snapshots of the swarm avoiding obstacles to reach a goal without external localization or communication. More details can be found in the attached video at https://youtu.be/U4i3Spisugg.}
    \label{fig:trajectory}
    \vspace{-12pt}
\end{figure}

However, real-world deployment of UAV swarms faces critical challenges in communication-constrained environments. Inter-UAV communication can become unavailable or unreliable due to electronic jamming, signal interference, or natural disasters~\cite{coppola2020survey}. Moreover, even when communication is available, the bandwidth requirements and latency constraints for large-scale swarms can limit scalability and real-time coordination~\cite{chen2020toward}. In such scenarios, communication-dependent methods become ineffective, necessitating robust communication-free collective navigation capabilities that rely solely on onboard perception and local decision-making.

Biological studies have provided significant insights into communication-free collective navigation. Berdahl \textit{et al.}~\cite{berdahl2018collective} identified five key mechanisms in animal collective behavior: the many wrongs principle (averaging individual errors), leadership (guidance by knowledgeable individuals), emergent sensing (pooling sensory information), social learning, and collective learning. Particularly relevant to this work, Couzin \textit{et al.}~\cite{couzin2005effective} demonstrated that effective group navigation can emerge from simple local interactions, where a small number of informed individuals can guide entire groups toward a destination without explicit communication or identification. This information asymmetry offers significant practical advantages for UAV swarms: it eliminates the need for mission broadcasting to all agents, reduces computational overhead, and enables scenarios where mission updates can only reach a subset of accessible UAVs. In communication-denied environments, where distributing goal information to the entire swarm may be infeasible, this approach becomes essential. 

This biological phenomenon directly inspires the implicit leader-follower framework proposed in this paper. In this framework, follower UAVs navigate without goal information or explicit leader identification. Instead, followers learn to maintain cohesion with neighboring UAVs through local perception. As the informed leader navigates toward the goal, the entire swarm naturally moves in that direction through the followers' learned cohesion behavior, without requiring explicit coordination or mission broadcasting. This emergent coordination eliminates the need for leader identification or goal information while preserving robust collective navigation.

\subsection{Related Works}

\subsubsection{Communication-Dependent Swarm Navigation}

Current UAV swarm implementations for collective navigation are broadly categorized into centralized and distributed approaches based on coordination architectures. Centralized approaches rely on external infrastructure for global coordination and precise state estimation. Representative systems include motion capture-based control such as Crazyswarm~\cite{preiss2017crazyswarm} and nonlinear model predictive control~\cite{soria2021predictive}, which achieve high-precision indoor navigation in obstacle-known environments by leveraging ground control stations for agent tracking. While demonstrating excellent coordination accuracy, their infrastructure dependence limits applicability to controlled indoor settings and causes network collapse upon infrastructure failure.

Distributed methods achieve coordination through decentralized decision-making without centralized control, typically requiring inter-agent communication for state or trajectory sharing. MADER~\cite{tordesillas2021mader} segments trajectory planning into perception, polyhedral representation, and optimization for multi-agent coordination. EGO-Swarm~\cite{zhou2021ego} employs optimization-based methods partitioned into mapping, planning, and control modules, enabling autonomous flight in cluttered forests through trajectory information sharing. Zhou \textit{et al.}~\cite{zhou2022swarm} introduce a spatial-temporal optimization framework for fully autonomous UAV swarms navigating dense, unknown environments without external facilities. V\'{a}s\'{a}rhelyi \textit{et al.}~\cite{vasarhelyi2018optimized} demonstrate evolutionary optimization for outdoor flocking using GNSS localization and wireless inter-agent communication. Recent learning-based approaches include deep reinforcement learning (DRL) for collision avoidance in fixed-wing UAVs~\cite{yan2023collision} and end-to-end DRL for quadrotor swarms mapping local observations to motor thrusts, enabling zero-shot transfer to real robots in obstacle-dense environments~\cite{huang2024collision}. However, these distributed methods fundamentally rely on continuous inter-agent communication for coordination—whether for trajectory sharing or state information exchange. Additionally, they typically assume either global mission knowledge across all agents or the ability to explicitly identify and communicate with informed leaders, limiting their applicability when communication is infeasible.

\subsubsection{Perception Strategies for Communication-Free Swarm Navigation}

To enable operation when communication is unavailable, communication-free approaches employ onboard sensor-based perception for surrounding UAV detection and autonomous coordination, eliminating inter-agent communication dependency. Vision-based approaches have been extensively explored through diverse methodologies. Saska \textit{et al.}~\cite{saska2017system} achieved relative localization using monocular cameras with circular patterns in GNSS-denied environments. Schilling \textit{et al.}~\cite{schilling2019learning} utilized six omnidirectional cameras without position sharing or visual markers, with subsequent work~\cite{schilling2021vision} achieving markerless outdoor flocking with three UAVs. Ahmad \textit{et al.}~\cite{ahmad2022pacnav} validated PACNav using ultraviolet-based relative sensing for navigation with four UAVs in natural forests. Wang \textit{et al.}~\cite{wang2023collective} implemented a bio-inspired visual projection field (VPF) approach with six UAVs in controlled indoor environments.

However, vision-based systems face limitations. First, they suffer from illumination dependency, with performance degrading under varying lighting conditions including sunlight, shadows, or darkness. Second, monocular vision requires additional processing or motion to resolve scale ambiguity in depth estimation. Third, omnidirectional perception requires multiple cameras, critically increasing system complexity through increased onboard computational demands for processing multiple video streams and complex inter-camera calibration and synchronization. These factors become particularly prohibitive for small UAV platforms with limited computation budgets. 

LiDAR-based approaches provide promising alternatives that directly address these vision-based limitations. Unlike vision systems, LiDAR offers illumination-invariant perception, maintaining consistent performance across diverse lighting conditions. The direct time-of-flight measurements eliminate depth ambiguity inherent in monocular vision, while a single 360-degree scanning LiDAR achieves omnidirectional awareness without the computational overhead, multi-stream processing, and complex calibration requirements of multi-camera systems. These characteristics make LiDAR particularly suitable for resource-constrained UAV platforms. Recent work has begun exploring LiDAR for swarm coordination. Swarm-LIO2~\cite{zhu2024swarm} demonstrates a fully decentralized LiDAR-inertial state estimation system for aerial swarms, achieving robust neighbor detection through reflective tape that produces distinctive high-intensity returns in LiDAR reflectivity measurements. The system has been validated with five UAVs in real-world experiments, demonstrating the practicality of LiDAR-based perception for swarm coordination.

\subsubsection{Control Strategies for Communication-Free Swarm Navigation}

While communication-free perception enables neighbor detection without information exchange, achieving truly autonomous coordination requires control strategies that make navigation decisions based purely on local sensory observations, without communicated state information. This includes challenges of real-time occlusion handling, limited field-of-view (FOV), complex swarm interactions, and obstacle avoidance using only onboard sensor data. Traditional heuristic methods, such as Reynolds' principles~\cite{reynolds1987flocks} and potential field-based obstacle avoidance~\cite{olfati2006flocking}, provide stable flocking without communication but exhibit limitations. Manually designing fixed rules that generalize across diverse environmental conditions—including varying obstacle densities, swarm sizes, and spatial constraints—is extremely challenging. These rule-based approaches often suffer from local minima where conflicting objectives trap agents in suboptimal behaviors~\cite{olfati2006flocking} and cannot adaptively adjust to dynamic swarm configurations~\cite{schilling2021vision,ahmad2022pacnav,wang2023collective}, often resulting in overly conservative maneuvers.

Learning-based approaches emerged to address these limitations, enabling agents to learn adaptive control policies for balancing flocking and obstacle avoidance in communication-free settings. Imitation learning trains UAV policies using expert demonstrations, partially mitigating heuristic rule rigidity, but is constrained by the quality and coverage of expert data~\cite{schilling2019learning,wan2024distributed}. Acquiring high-quality demonstration data for communication-free coordination is inherently challenging, requiring experts to manually demonstrate complex behaviors under diverse conditions with partial observability.

DRL offers a powerful alternative, overcoming both heuristic rule rigidity and imitation learning's data dependency by enabling agents to learn policies through direct environmental interaction. DRL can automatically discover strategies to balance multiple competing objectives—maintaining cohesion, avoiding collisions, and handling perception uncertainties—through reward-driven optimization, without requiring explicit rule design or expert demonstrations. Huang \textit{et al.}~\cite{huang2022vision} proposed vision-based decentralized collision avoidance using depth images and DRL for multi-UAV navigation without inter-UAV communication, but validated only in obstacle-free simulation environments. Bai \textit{et al.}~\cite{bai2023learning} extended this for communication-denied environments with limited visual fields, but validated in simulations with simplified dynamics and pre-known obstacles, limiting real-world applicability assessment. As existing learning-based approaches are mostly validated in simulations, a critical gap in real-world validation remains, which this study addresses.

\subsection{Contributions}
To overcome these limitations, we propose a novel LiDAR-based collective navigation system for robust operation in GNSS-denied and communication-denied environments. Building upon~\cite{zhu2024swarm}, we eliminate all communication dependencies by enabling coordination through purely local sensing. Each UAV uses a single LiDAR sensor for 360-degree neighbor detection, removing the need for external localization or inter-agent communication. We employ an implicit leader-follower architecture where only the leader possesses goal information, while followers coordinate through learned reactive behaviors. Follower control policies are trained via DRL using proximal policy optimization (PPO)~\cite{schulman2017proximal} to learn flocking and obstacle avoidance solely from local LiDAR observations. As an on-policy algorithm, PPO enables synchronous data collection from multiple UAVs and efficient batch updates in GPU-accelerated parallel environments. The DRL training enables followers to autonomously discover control strategies that balance cohesion with neighbors and collision avoidance with obstacles. Through this learned flocking behavior, the swarm collectively navigates to the destination as the leader moves toward the goal, without requiring explicit leader identification or goal information. The proposed system is validated through extensive simulations and real-world experiments involving five UAVs. A sample result of the real-world experiments is depicted in Fig.~\ref{fig:trajectory}. To the best of our knowledge, this is the first LiDAR-based collective navigation system for a swarm of UAVs using DRL that operates without any information exchange and has been validated in real-world deployments. The key contributions are:

\begin{enumerate}
    \renewcommand{\labelenumi}{\roman{enumi})}
    \item A fully communication-free, LiDAR-based perception framework for neighbor detection and tracking to achieve robust UAV swarm coordination;
    \item A DRL-based control policy that enables implicit leader-follower coordination by learning to balance flocking and obstacle avoidance under realistic perception constraints; and
    \item Comprehensive validation via extensive simulations and real-world experiments with five UAVs in diverse indoor and outdoor environments.
\end{enumerate}

\section{Problem Formulation}
\label{sec:problem_formulation}

\subsection{Problem Definition}
\begin{figure}[t!]
    \centering
    \includegraphics[width=\columnwidth]{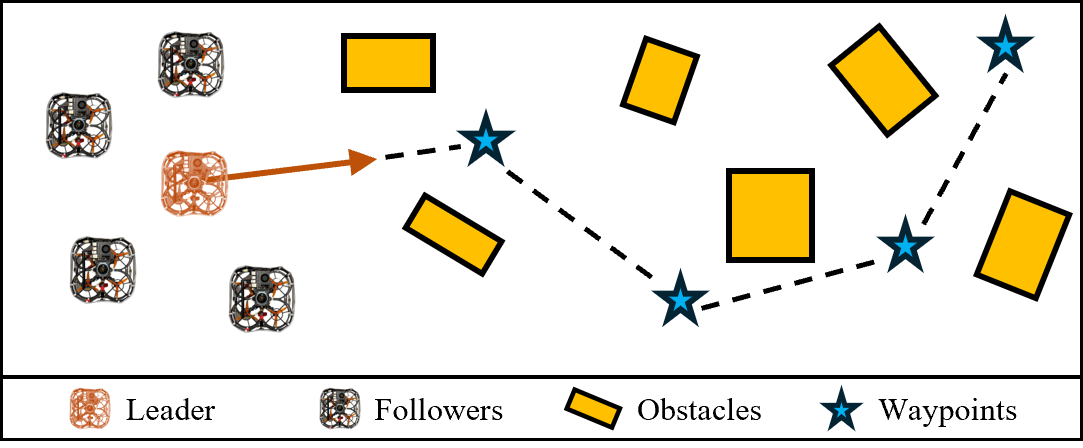}
    \vspace{-20pt}
    \caption{Scenario of the communication-free collective navigation problem. A leader visits a known waypoint sequence, while followers, unaware of the goal or leader's identity, use only local perception for flocking and obstacle avoidance to indirectly follow the leader.}
    \label{fig:scenario}
    \vspace{-12pt}
\end{figure}

This study addresses the collective navigation of a swarm of $N$ UAVs toward a specific destination in a GNSS-denied and communication-constrained environment. As illustrated in Fig.~\ref{fig:scenario}, the swarm consists of one informed leader UAV, which possesses destination information via pre-loaded waypoints, and $N-$1 uninformed follower UAVs without such information. We assume a complete absence of information exchange, meaning each UAV cannot explicitly distinguish whether other UAVs are leaders or followers, requiring an implicit leader-follower framework. Each follower must utilize its onboard LiDAR sensors to perceive its surroundings in real-time, making autonomous decisions based only on locally observed information. The core challenge is for the swarm to maintain cohesion and effectively avoid obstacles, moving collectively toward the destination without dispersing, while addressing perceptual challenges such as occlusion and limited FOV.

\subsection{System Models}
The state of each UAV $i$ is its local position $\mathbf{p}^i_t \in \mathbb{R}^3$, velocity $\mathbf{v}^i_t \in \mathbb{R}^3$, and orientation (quaternion $\mathbf{q}^i_t \in \mathbb{R}^4$) at time $t$. Each UAV $i$ has a LiDAR sensor providing a raw point cloud $\mathbf{z}^i_{t}$ of its surroundings, which feeds into a perception module (Sec.~\ref{sec:perception}). From this data, UAV $i$'s state $(\mathbf{p}^i_t, \mathbf{v}^i_t, \mathbf{q}^i_t)$ is estimated by LiDAR-inertial odometry (LIO). Concurrently, the perception module processes $\mathbf{z}^i_{t}$ to detect and track neighbors $\mathcal{N}^i_{t}$ and obstacles $\mathcal{O}^i_{t}$.

While all UAVs share the same kinematic model for position updates,
\begin{equation*}
    \mathbf{p}^i_{t+1} = \mathbf{p}^i_{t} + \mathbf{v}^i_{t}\Delta t,
\end{equation*}
their control policies for determining velocity commands differ by role.

The leader's velocity is determined by a local planner $L$ using its estimated state, perceived obstacles, and predefined waypoints $W$:
\begin{equation*}
    \mathbf{v}^l_{t+1} = L(\mathbf{p}^l_{t}, \mathbf{q}^l_{t}, \mathcal{O}^l_{t}, W).
\end{equation*}
Each follower learns a control policy $\pi$, detailed in Sec.~\ref{sec:DRL}, that generates velocity commands based only on its estimated state and local perception of neighbors and obstacles:
\begin{equation*}
    \mathbf{v}^{f}_{t+1} = \pi(\mathbf{v}^{f}_{t}, \mathbf{q}^{f}_{t}, \mathcal{N}^{f}_{t}, \mathcal{O}^{f}_{t}).
\end{equation*}

The follower's policy $\pi$ intentionally excludes position $\mathbf{p}^{f}_{t}$ as an input, even though it is available from LIO. This design is critical for learning a generalizable policy. Including $\mathbf{p}^{f}_{t}$ could cause the policy to learn correlations between coordinates within the locally generated map and required control actions. Such map-dependent policies would fail in novel environments with different trajectories and layouts. By restricting input to the UAV's kinematic state—velocity $\mathbf{v}^{f}_t$ and orientation $\mathbf{q}^{f}_t$—along with local perception of neighbors $\mathcal{N}^f_{t}$ and obstacles $\mathcal{O}^f_{t}$, the policy learns a robust, egocentric strategy based on immediate kinematic state and surroundings rather than position relative to an arbitrary starting point, significantly enhancing transferability.

\subsection{Flocking Considerations for Implicit Following}
Flocking refers to coordinated collective motion emerging when individual agents follow simple local interaction rules~\cite{reynolds1987flocks}. Traditional flocking behaviors consist of three fundamental rules: cohesion for staying close to neighboring UAVs, separation for maintaining safe distances to avoid collisions with neighbors, and alignment for matching velocity directions with neighbors.

However, we deliberately exclude the alignment term from our flocking behavior. This design choice is crucial for our communication-free implicit leader-follower scenario, where followers cannot identify the leader and the leader may exhibit diverse motion patterns—including hovering, directional changes, or backward movement—depending on mission objectives and obstacle configurations. Incorporating velocity alignment would cause followers to align with the average velocity direction of all perceived neighbors rather than specifically following the leader's directional changes. Since followers cannot distinguish the leader, when the leader suddenly changes direction to navigate toward a new location, followers maintaining alignment with neighbors' average velocity would dilute the leader's influence, potentially causing loss of the leader and mission failure. Therefore, our flocking behavior incorporates only cohesion and separation, allowing followers to navigate collectively while being responsive to the leader's positional changes rather than being constrained by the swarm's average velocity direction.

\subsection{Objectives}
The primary objective of this study is to develop a robust control policy, $\pi$, trained via DRL. Using only onboard LiDAR data, this policy must enable followers to learn a balance between two types of local behaviors: flocking (cohesion and separation) and obstacle avoidance. The successful execution of these behaviors results in the emergent behavior of implicit leader-following, where the swarm naturally moves toward the destination as followers maintain positional cohesion with neighbors, which in turn are influenced by the leader's goal-directed motion. Ultimately, the system aims to achieve collective movement in a fully communication-free manner.

\section{Methodology}

\begin{figure*}[t!]
    \centering
    \includegraphics[width=0.8\textwidth]{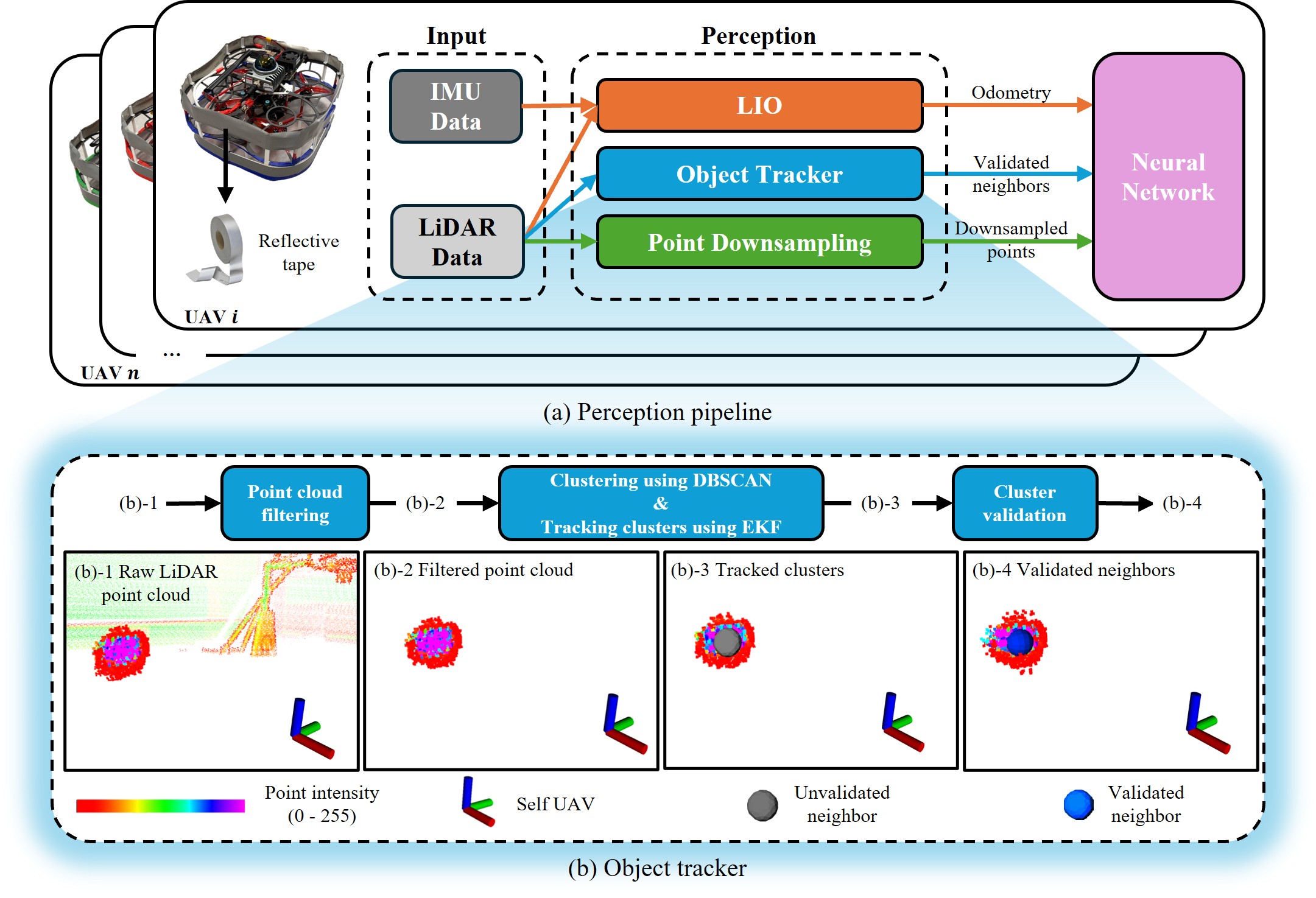}
    \vspace{-10pt}
    \caption{Overview of the onboard perception system. (a) The perception pipeline. (b) The object tracker filters raw LiDAR points, clusters them using DBSCAN, tracks them with an EKF, and validates them based on temporal consistency.}
    \label{fig:stack}
    \vspace{-12pt}
\end{figure*}

To address the problem defined in Sec.~\ref{sec:problem_formulation}, the proposed system is built upon a decentralized control architecture where each UAV independently perceives its environment and makes decisions. This architecture is composed of two core modules: a LiDAR-based perception system (Fig.~\ref{fig:stack}) that detects neighboring UAVs and obstacles in real-time, and a DRL control policy (Fig.~\ref{fig:drl}) that learns to balance the competing demands of flocking and obstacle avoidance. The perception system generates local observations that are fed as an input to the control policy, which in turn outputs velocity commands. The integration of these modules enables the emergent behavior of stable collective navigation, allowing the entire swarm to reach a target destination known only to the leader, without explicit communication or reliance on an external positioning system.

\subsection{LiDAR-Based Perception System}
\label{sec:perception}

The proposed LiDAR-based perception pipeline consists of three main components: (1) a LIO module for ego-state estimation, (2) an object tracker module for neighbor detection and tracking, and (3) a point downsampling module that transforms the raw point cloud into low-dimensional features for computational efficiency. The resulting perception outputs are fed into the neural network. Each UAV is equipped with reflective tape to produce high-intensity LiDAR returns, crucial for detection. Based on the high-intensity returns, the object tracker operates in three stages with all geometry computed in the local map frame: filtering, clustering/tracking, and validation.

\subsubsection{Point Cloud Filtering}
The first stage filters the raw point cloud of each UAV $i$, $\mathbf{z}^i_{t}$, as shown in Fig.~\ref{fig:stack}(b)-1. To improve point density, we stack the most recent $B$ point clouds. Each point ${^b}\mathbf{p}$ is transformed into the map frame using the transformation ${^m}\mathbf{p} = T_{m\leftarrow b}\,{^b}\mathbf{p}$, where $T_{m\leftarrow b}$ is obtained from the LIO module and subsequently gated by its Euclidean distance to the UAV position, retaining only points within range $[d_{\min}, d_{\max}]$.

The resulting distance-gated points, $\mathcal{P}_d$, are filtered based on two criteria. First, we identify high-intensity points $\mathcal{P}_{\text{high}}$. Since reflective tape provides high-intensity returns, points with intensity $I$ greater than or equal to threshold $I_{\text{high}}$ are utilized as key seeds to detect new objects.
\begin{equation*}
    \mathcal{P}_{\text{high}} = \{\,\mathbf{p} \in \mathcal{P}_d \;|\; I \ge I_{\text{high}} \,\}.
\end{equation*}

Second, we retain region of interest (ROI) points $\mathcal{P}_{\text{roi}}$ to maintain tracks during temporary occlusions. From remaining low-intensity points $\mathcal{P}_{\text{low}}$, we select those within radius $r_{\text{roi}}$ of existing track centroids ${^m}\hat{\mathbf{c}}^j$. 
\begin{equation*}
    \mathcal{P}_{\text{roi}} = \{\,\mathbf{p} \in \mathcal{P}_{\text{low}} \;|\; \min_j \| {^m}\mathbf{p} - {^m}\hat{\mathbf{c}}^j \|_2 \le r_{\text{roi}} \}.
\end{equation*}
Finally, the union $\mathcal{P}_{\text{filtered}}=\mathcal{P}_{\text{high}}\cup \mathcal{P}_{\text{roi}}$ is passed to the next stage as shown in Fig.~\ref{fig:stack}(b)-2.

\subsubsection{Clustering and Tracking}
The filtered point set $\mathcal{P}_{\text{filtered}}$ is grouped into individual clusters $C^k$ using the density-based spatial clustering of applications with noise (DBSCAN) algorithm~\cite{ester1996density}, which forms clusters based on a distance threshold $\varepsilon$ defining the maximum allowable distance between points and a minimum number of points $n_\text{min}$. Since point density decreases with distance, we apply distance-adaptive validation where the minimum point threshold scales with range. Each cluster is tracked using the extended Kalman filter (EKF)~\cite{kalman1960new} with a constant velocity model to estimate smooth motion trajectories of neighboring UAVs, as depicted in Fig.~\ref{fig:stack}(b)-3. Data association matches each valid cluster to the nearest existing track if their distance is below threshold $d_{\text{match}}$. An associated track is updated via EKF, and a new track is created for a new cluster. A track is deactivated if unobserved for timeout $t_{\text{inactive}}$.

\begin{table}[t!]
\caption{Parameters for the LiDAR-Based Perception System}
\label{tab:perception_params}
\resizebox{\columnwidth}{!}{%
\centering
\begin{tabular}{ccp{6cm}}
\hline
\textbf{Parameter} & \textbf{Value} & \textbf{Description} \\
\hline
\multicolumn{3}{l}{\textit{Point Cloud Filtering}} \\
$B$ & 2 & Number of stacked point clouds \\
$d_{\min}$ & 0.05m & Minimum distance for point gating \\
$d_{\max}$ & 10.0m & Maximum distance for point gating \\
$I_{\text{high}}$ & 170 & Threshold for LiDAR intensity (0–255) for highly reflective points \\
$r_{\text{roi}}$ & 0.3m & ROI radius around existing tracks \\
\hline
\multicolumn{3}{l}{\textit{Clustering and Tracking}} \\
$\varepsilon$ & 0.1m & Distance threshold for DBSCAN clustering \\
$n_\text{min}$ & $8B$ & Minimum points per cluster (scaled by stacked frames) \\
$d_{\text{match}}$ & 0.2m & Association threshold for matching clusters to tracks \\
$t_{\text{inactive}}$ & 0.5s & Timeout for deactivating unobserved tracks \\
\hline
\multicolumn{3}{l}{\textit{Cluster Validation}} \\
$\rho_{\text{high}}$ & 0.05 & High-intensity ratio threshold for validation \\
$\tau_{\text{on}}$ & 0.01s & Continuous duration for validation \\
\hline
\end{tabular}
}
\vspace{-12pt}
\end{table}

\subsubsection{Cluster Validation}
The final stage validates if a tracked object is a reliable neighbor UAV as shown in Fig.~\ref{fig:stack}(b)-4. A track is confirmed as a reliable neighbor if the ratio of high-intensity points within its cluster exceeds threshold $\rho_{\text{high}}$ for continuous duration $\tau_{\text{on}}$. The set of all validated neighbors at time $t$ constitutes the perception output $\mathcal{N}_{t}$, used by the control policy.

\subsection{Deep Reinforcement Learning Framework}
\label{sec:DRL}

\begin{figure*}[t!]
    \centering
    \includegraphics[width=0.8\textwidth]{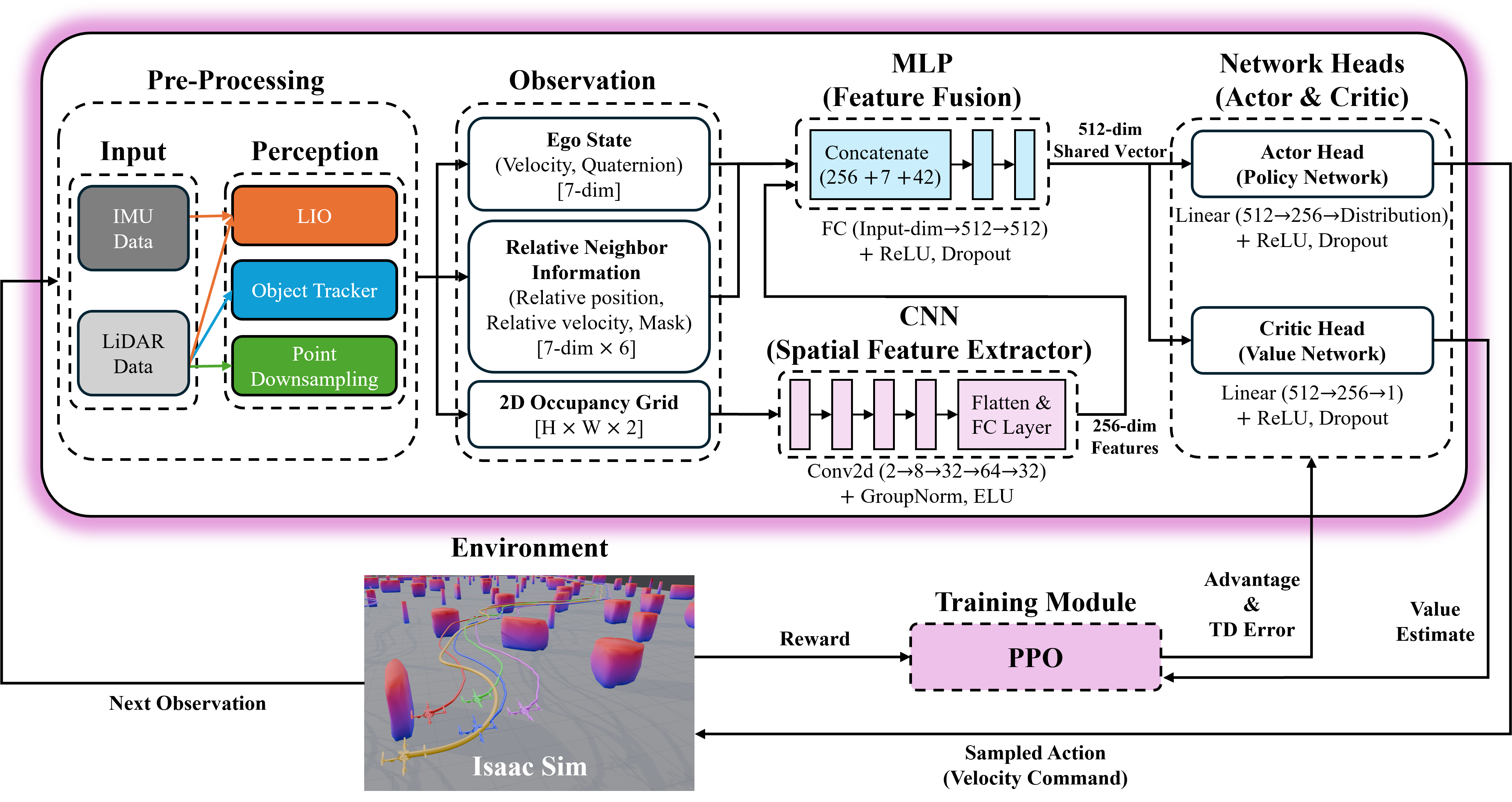}
    \vspace{-10pt}
    \caption{The proposed DRL architecture. An encoder processes observations into a latent vector, and actor and critic heads use it to determine the policy and estimate value.}
    \label{fig:drl}
    \vspace{-12pt}
\end{figure*}

The follower UAVs learn a decentralized control policy using a DRL framework, as illustrated in Fig.~\ref{fig:drl}. We model the follower's control problem as a partially observable Markov decision process (POMDP), formally defined by the 7-tuple $(\mathcal{S}, \mathcal{A}, \mathcal{T}, \mathcal{R}, \Omega, \mathcal{O}, \gamma)$, where $\mathcal{S}$ is the state space, $\mathcal{A}$ is the action space, $\mathcal{T}: \mathcal{S} \times \mathcal{A} \times \mathcal{S} \rightarrow [0,1]$ is the state transition function, $\mathcal{R}: \mathcal{S} \times \mathcal{A} \rightarrow \mathbb{R}$ is the reward function, $\Omega$ is the observation space, $\mathcal{O}: \mathcal{S} \rightarrow \Omega$ is the observation function, and $\gamma \in [0,1]$ is the discount factor. In our model-free DRL approach, the agent does not explicitly know $\mathcal{S}$, $\mathcal{T}$, or $\mathcal{O}$. Instead, it learns an optimal policy $\pi: \Omega \rightarrow \mathcal{A}$ through direct interaction with the environment, relying on observations $o_t \in \Omega$, actions $a_t \in \mathcal{A}$, and rewards $r_t \in \mathbb{R}$ sampled from $\mathcal{R}$.

\subsubsection{Observation and Action Spaces}
The agent's observation $o_t \in \Omega$ at time step $t$ is a composite input vector designed for decentralized control, comprising: (1) a 7-D ego-state vector (3-D velocity and 4-D quaternion) for self-motion awareness, (2) a 42-D vector encoding the relative states of up to six neighbors, and (3) a two-channel $H \times W$ occupancy grid derived from LiDAR data for spatial understanding of obstacles, where $H=72$ and $W=12$ correspond to the horizontal and vertical resolution of the downsampled point cloud.

For neighbor representation, we select up to six nearest neighbors visible within the LiDAR's FOV, accounting for occlusions that may temporarily hide agents. Each neighbor is encoded as a 7-D vector containing relative position, relative velocity, and a binary mask. The use of a fixed number of observable neighbors is a topological approach inspired by robust collective behaviors in animal swarms~\cite{ballerini2008interaction}. The selection of six neighbors balances performance and computational load, empirically verified in Sec.~\ref{sec:ablation}. When fewer than six neighbors are detected due to occlusions or limited vertical FOV, zero-padding and binary masks maintain a fixed input size while indicating the presence or absence of valid neighbor information.

The occupancy grid's first channel contains proximity information, representing distance to the nearest obstacle in each cell, while the second channel provides a binary mask indicating cell occupancy. Based on this observation, the agent generates an action $a_t \in \mathcal{A}$, a continuous 3-D vector representing the desired velocity command $\mathbf{v} \in \mathbb{R}^3$.

\subsubsection{Reward Function}
The reward function $\mathcal{R}$ is a multi-objective sum designed to encourage complex emergent behaviors, with total reward:
\begin{equation*}
    r_{\text{total}} = r_{\text{flock}} + r_{\text{obstacle}} + r_{\text{stable}} + r_{\text{perception}} + r_{\text{collision}}.
\end{equation*}

The flocking reward balances separation from nearby agents and cohesion with the flock's center of mass:
\begin{equation*}
    r_{\text{flock}} = w_{\text{flock}} (r_{\text{separation}} + r_{\text{cohesion}}),
\end{equation*}
where $r_{\text{separation}}$ penalizes violating a safety distance, and $r_{\text{cohesion}}$ penalizes straying from the flock's center.
\begin{align*}
    r_{\text{separation}} &= -\sum_{j \in \mathcal{N}^i} \left( \frac{d_{\text{sep}} - \|\mathbf{p}^i - \mathbf{p}^j\|_2}{d_{\text{sep}} - 2 r_{\text{uav}}} \right) \mathbb{I}_{\|\mathbf{p}^i - \mathbf{p}^j\|_2 < d_{\text{sep}}}, \\ r_{\text{cohesion}} &= - \left( \|\mathbf{p}^i - \mathbf{p}_{\text{com}}\|_2 - d_{\text{coh}} \right) \mathbb{I}_{\|\mathbf{p}^i - \mathbf{p}_{\text{com}}\|_2 > d_{\text{coh}}}.
\end{align*}
Here, $\mathbf{p}^i$ is the UAV's position, $\mathbf{p}^j$ is the neighbor's position, $\mathbf{p}_{\text{com}}$ is the center of mass, $d_{\text{sep}}$ and $d_{\text{coh}}$ are distance thresholds, $r_{\text{uav}}$ is the UAV radius, and $\mathbb{I}$ is the indicator function.

The obstacle avoidance reward penalizes obstacle proximity and approach:
\begin{equation*}
    r_{\text{obstacle}} = w_{\text{obstacle}} (r_{\text{proximity}} + r_{\text{direction}}),
\end{equation*}
where $r_{\text{proximity}}$ penalizes distance to the nearest obstacle, and $r_{\text{direction}}$ penalizes moving towards obstacles.
\begin{align*}
    r_{\text{proximity}} &= -\left(\frac{d_{\text{prox}} - \min_k(d^k_{\text{obs}})}{d_{\text{prox}} - r_{\text{uav}}}\right)^4 \mathbb{I}_{\min_k d^k_{\text{obs}} < d_{\text{prox}}}, \\
    r_{\text{direction}} &= -\sum_k \max(0, d_{\text{prox}} - d^k_{\text{obs}}) \|\mathbf{v}^i\|_2 \mathbb{I}_{|\theta^k| < \theta_{\text{threshold}}}.
\end{align*}
Here, $k$ indexes the individual obstacle points observed by the LiDAR; $d^k_{\text{obs}}$ is the distance to the $k$-th point, and $\theta^k$ is the angle between the UAV's velocity and the ray to the $k$-th obstacle point. $d_{\text{prox}}$ and $\theta_{\text{threshold}}$ are defined as the proximity and angle thresholds used for obstacle avoidance.

The stable flight reward is:
\begin{equation*}
    r_{\text{stable}} = w_{\text{stable}} (r_{\text{altitude}} + r_{\text{attitude}}),
\end{equation*}
including $r_{\text{altitude}}$ for altitude maintenance and $r_{\text{attitude}}$ for attitude stability.
\begin{align*}
    r_{\text{altitude}} &= \exp\left(-\left(\frac{h^i - h^l}{\alpha}\right)^2\right), \\
    r_{\text{attitude}} &= \exp\left(-\left(\frac{u_z - 1}{\beta}\right)^2\right).
\end{align*}
Here, $h^i$ is the UAV's altitude, $h^l$ is the leader's altitude, and $u_z$ is the z-component of the UAV's up-vector. Parameters $\alpha$ and $\beta$ control sensitivity.

\begin{figure}[t!]
    \centering
    \includegraphics[width=\columnwidth]{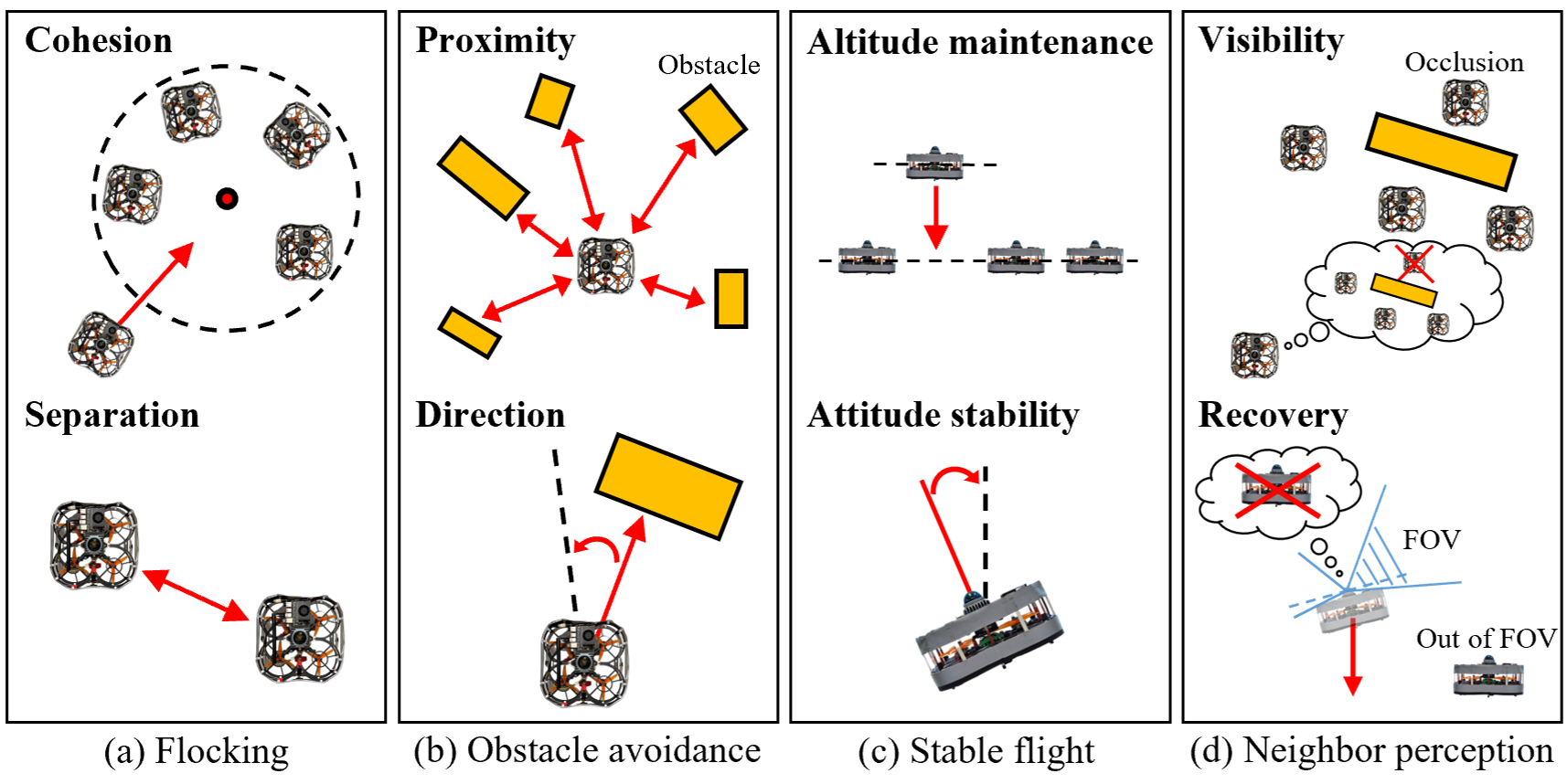}
    \vspace{-20pt}
    \caption{Illustration of reward function components for learning communication-free collective navigation. (a) Flocking: cohesion encourages staying close to neighbors, while separation maintains safe distances. (b) Obstacle avoidance: proximity penalizes closeness to obstacles, and direction penalizes movement toward obstacles. (c) Stable flight: altitude maintenance encourages following the leader's altitude, and attitude stability promotes upright orientation. (d) Neighbor perception: visibility rewards keeping neighbors within FOV, recovery triggers descent when all neighbors are lost.}
    \label{fig:reward_components}
    \vspace{-12pt}
\end{figure}

The neighbor perception reward encourages maintaining awareness of neighboring UAVs:
\begin{equation*}
    r_{\text{perception}} = w_{\text{perception}} (r_{\text{visibility}} + r_{\text{recovery}}),
\end{equation*}
composed of $r_{\text{visibility}}$ for keeping neighbors within FOV by minimizing occlusion and $r_{\text{recovery}}$ for descent when no neighbors are detected. This descent acts as a recovery maneuver to re-acquire neighbors that may have fallen out of the limited vertical FOV.
\begin{align*}
    r_{\text{visibility}} &= \frac{|\mathcal{N}^i_{\text{perceived}}|}{|\mathcal{N}^i|}, \\
    r_{\text{recovery}} &= -|h^i - h_{\text{recovery}}| \mathbb{I}_{\mathcal{N}^i_{\text{perceived}}=0}.
\end{align*}
Here, $\mathcal{N}^i_{\text{perceived}}$ is the set of neighbors detected by LiDAR accounting for occlusions and limited FOV, and $h_{\text{recovery}}$ is the target descent altitude.

Finally, the collision penalty is a sparse negative reward applied upon any collision event and triggers episode termination:
\begin{equation*}
    r_{\text{collision}} = -10 \mathbb{I}_{\text{collision}}.
\end{equation*}
A collision event is triggered if the distance between centers of any two UAVs becomes less than $2 r_{\text{uav}}$, or the distance between a UAV's center and an obstacle point becomes less than $r_{\text{uav}}$.

\begin{table}[t!]
\caption{Parameters for the Reward Function}
\resizebox{\columnwidth}{!}{%
\label{tab:reward_params}
\centering
\begin{tabular}{ccp{6cm}}
\hline
\textbf{Parameter} & \textbf{Value} & \textbf{Description} \\
\hline
\multicolumn{3}{l}{\textit{Reward Weights}} \\
$w_{\text{flock}}$ & 1.5 & Weight for the flocking reward \\
$w_{\text{obstacle}}$ & 2.0 & Weight for the obstacle avoidance reward \\
$w_{\text{stable}}$ & 1.0 & Weight for the stable flight reward \\
$w_{\text{aware}}$ & 1.0 & Weight for the situational awareness reward \\
\hline
\multicolumn{3}{l}{\textit{Flocking}} \\
$d_{\text{sep}}$ & 1.6m & Safety distance for separation \\
$d_{\text{coh}}$ & 2.0m & Cohesion distance threshold from center of mass \\
$r_{\text{uav}}$ & 0.2m & Radius of the UAV \\
\hline
\multicolumn{3}{l}{\textit{Obstacle Avoidance}} \\
$d_{\text{prox}}$ & 3.0m & Proximity threshold for obstacles \\
$\theta_{\text{threshold}}$ & $20^\circ$  & Angle threshold for direction penalty \\
\hline
\multicolumn{3}{l}{\textit{Stable Flight}} \\
$\alpha$ & 0.1 & Scaling parameter for altitude reward \\
$\beta$ & 0.1 & Scaling parameter for attitude reward \\
\hline
\multicolumn{3}{l}{\textit{Situational Awareness}} \\
$h_{\text{recovery}}$ & 1.0 m & Target altitude for emergency descent \\
\hline
\end{tabular}
}
\vspace{-12pt}
\end{table}

\subsubsection{Network Architecture and Training}

Our control policy is implemented as an actor-critic network, illustrated in Fig.~\ref{fig:drl}. The network consists of a shared encoder processing observations, followed by actor and critic heads. The entire policy is trained end-to-end using the PPO algorithm~\cite{schulman2017proximal}.

The two-channel $72 \times 12$ LiDAR occupancy grid is processed by a convolutional neural network (CNN). The resulting feature map is flattened and passed through a fully-connected layer to produce a 256-D feature vector, which is concatenated with the 7-D ego-state vector and 42-D neighbor-state vector. The combined feature vector is passed through a multi-layer perceptron (MLP), outputting a final 512-D shared feature vector. The shared vector is consumed by two separate heads. The actor head is an MLP with a hidden layer of 256 units that outputs the mean and standard deviation for a continuous Gaussian policy. The critic head is a similar MLP outputting a single scalar representing the value function.

The training process consists of two core stages: advantage estimation, followed by policy and value function optimization. To evaluate how much better a specific action is compared with the average action in a given state, we use an advantage function, $\hat{A}_t$. We apply generalized advantage estimation (GAE)~\cite{schulman2015high}, which computes advantages based on the temporal difference (TD) error, $\delta_t$:
\begin{equation}
    \delta_t = R_t + \gamma V_\phi(o_{t+1}) - V_\phi(o_t),
\end{equation}
where $R_t$ is the reward, $V_\phi$ is the value estimated by the critic network, and $\gamma$ is the discount factor. GAE calculates the advantage as an exponentially-weighted sum of these TD errors:
\begin{equation}
    \hat{A}_t^{\text{GAE}(\gamma, \lambda)} = \sum_{l=0}^{T-t-1} (\gamma \lambda)^l \delta_{t+l},
\end{equation}
where $\lambda \in [0, 1]$ controls the bias-variance trade-off. The resulting advantages $\hat{A}_t$ are normalized to zero mean and unit variance before policy optimization.

The policy and value networks are optimized jointly by minimizing a total loss function:
\begin{equation}
    L(\theta,\phi) = \hat{\mathbb{E}}_t [ -L^{\text{CLIP}}(\theta) + c_1 L_{\text{VF}}(\phi) - c_2 H[\pi_\theta](o_t) ],
\end{equation}
where $c_1$ and $c_2$ are hyperparameters weighting the different terms. The primary component is the clipped surrogate objective:
\begin{equation}
    L^{\text{CLIP}}(\theta) = \hat{\mathbb{E}}_t \left[ \min \left( r_t(\theta) \hat{A}_t, \text{clip}(r_t(\theta), 1-\epsilon, 1+\epsilon) \hat{A}_t \right) \right].
\end{equation}
Here, $r_t(\theta) = \frac{\pi_\theta(a_t | o_t)}{\pi_{\theta_{\text{old}}}(a_t | o_t)}$ is the probability ratio between current and old policies. This clipping mechanism is the core of PPO's stability, discouraging overly large policy updates.

\begin{figure*}[t!]
    \centering
    \includegraphics[width=0.9\textwidth]{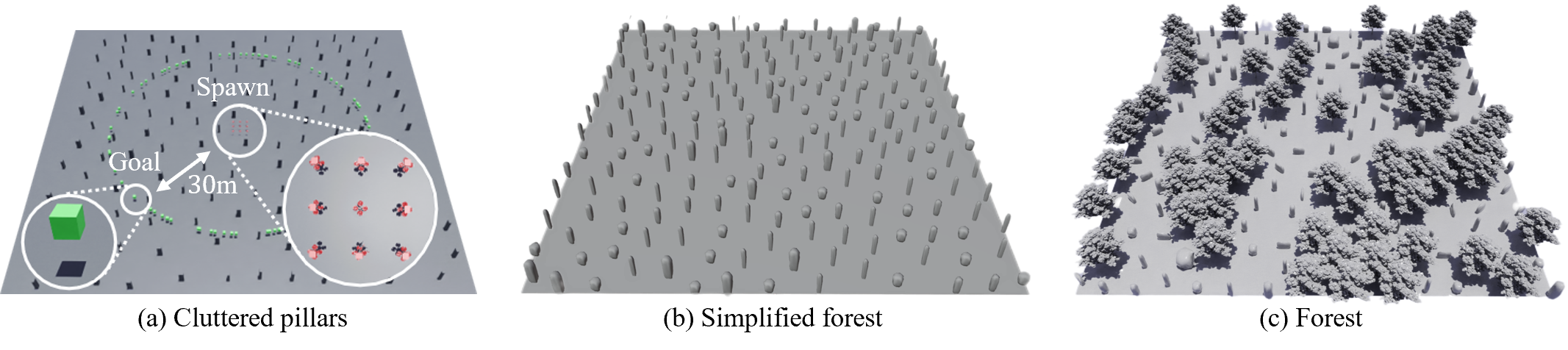}
    \vspace{-10pt}
    \caption{Simulation environments. (a) The cluttered training environment with spawn area at the map center and goal point randomly sampled on a circle of 30m radius centered at the spawn region. (b) and (c) show unseen test environments for evaluating generalization.}
    \label{fig:simulation}
    \vspace{-12pt}
\end{figure*}

The second component, the value function loss $L_{\text{VF}}$, reduces the variance of advantage estimates. The value network minimizes the error between its prediction $V_\phi(o_t)$ and the value target $V_t^{\text{targ}}$, computed using value estimates from the network used during data collection:
\begin{equation}
    V_t^{\text{targ}} = \hat{A}_t + V_{\phi_{\text{old}}}(o_t).
\end{equation}
This loss is the squared error: 
\begin{equation}
    L_{\text{VF}}(\phi) = (V_\phi(o_t) - V_t^{\text{targ}})^2.
\end{equation}
Finally, an entropy bonus $H[\pi_\theta](o_t)$ encourages exploration, where $H$ is the entropy of the policy's action distribution: 
\begin{equation}
    H[\pi_\theta](o_t) = \mathbb{E}_{a \sim \pi_\theta}[-\log \pi_\theta(a|o_t)].
\end{equation} 
Key hyperparameters are set to $\epsilon=0.1$, $c_1=1.0$, $c_2=0.001$, $\gamma=0.99$, $\lambda=0.95$, using the Adam optimizer with learning rates of $1\times10^{-3}$ for the encoder, actor, and critic.


\section{Simulations}
\subsection{Simulation Setup}
We trained and evaluated our DRL policy in Nvidia Isaac Sim using the OmniDrones~\cite{xu2024omnidrones} framework for GPU-accelerated parallel simulations. Each training simulation instance featured a swarm of five UAVs, comprising one leader and four followers, navigating an environment with randomly placed pillars as shown in Fig.~\ref{fig:simulation}(a).

\subsubsection{Environment Configuration} 
At the beginning of each training episode, the swarm was spawned in an obstacle-free region at the map center, with UAVs randomly positioned on a $3 \times 3$ grid with 1.6m spacing to ensure safe initial separation. The goal point was randomly sampled on a 30m radius circle centered at the spawn region. To enhance policy robustness and generalization, we randomized initial UAV orientations, the leader's speed, and goal location for each episode.

Training was conducted across 512 parallel environments for 500 million timesteps, leveraging GPU acceleration (Nvidia RTX 4090 and Intel i9-14900K) for efficient data collection and policy updates. Each episode terminated under four conditions: (1) collision between UAVs, (2) collision between a UAV and obstacle, (3) flying below minimum allowable altitude, and (4) flying above maximum allowable altitude, ensuring the policy learns to maintain swarm cohesion, avoid collisions, and preserve stable flight within operational constraints.

\subsubsection{Leader Navigation} 
The leader UAV utilized a planner combining the rapidly-exploring random tree (RRT)~\cite{lavalle1998rapidly} with the artificial potential field (APF)~\cite{khatib1986real} for real-time navigation. RRT generates a local path toward the waypoint to prevent APF from falling into local minima. Then, APF computes an attractive force toward the RRT-generated path direction and repulsive forces from nearby obstacles detected by the 360-degree LiDAR. The final velocity command is obtained by combining these APF forces, enabling the UAV to follow the RRT path while maintaining safe distances from obstacles. This integration allows smooth and safe navigation in unknown environments.

\subsubsection{Perception Modeling} 
While our physical system uses reflective tape for neighbor detection, in simulation we abstracted this process using the simulator's ground-truth data to identify neighbors. This abstraction allowed us to focus training on the control policy while incorporating key perceptual challenges based on our hardware's empirical characteristics: (1) estimation errors in relative neighbor states with Gaussian distributions (relative position: $\sigma_{\text{pos}} = 0.02$m, relative velocity: $\sigma_{\text{vel}} = 0.05$m/s), (2) limited FOV identical to the physical sensor ($-7^\circ$ to $52^\circ$ vertical), (3) occlusions calculated based on physical UAV dimensions, and (4) processing latency reflecting actual onboard computation. Latency was determined by measuring end-to-end delay on the onboard computer, from sensor data acquisition through our perception pipeline to policy input generation. To ensure temporal consistency, data synchronization introduces delays of 0.1s for ego state and LiDAR data for obstacles, and 0.2s for relative neighbor information.

\subsection{Evaluation Metrics}
To quantitatively evaluate the proposed policy, six metrics were defined: success rate (\textit{SR}), mission progress (\textit{MP}), flock radius (\textit{FR}), minimum separation (\textit{MS}), alignment (\textit{AL}), and minimum distance to obstacles (\textit{MDO}). All results represent mean and standard deviation from 100 independent trials per scenario.

\subsubsection{Success Rate (\textit{SR})}
The percentage of trials where all UAVs successfully followed the leader to the destination without collisions, serving as the primary indicator of mission completion. Evaluation terminates when: (1) UAV-UAV collision, (2) UAV-obstacle collision, (3) flying below/above altitude limits, or (4) complete leader loss from all followers' perception. \textit{SR} is defined as
$$
\text{\textit{SR}} = \frac{N_{\text{successful trials}}}{N_{\text{trials}}} \times 100.
$$

\subsubsection{Mission Progress (\textit{MP})}
The ratio of distance advanced toward the destination relative to total distance from start to destination, averaged over all trials. This evaluates navigation effectiveness and if collective behavior enables consistent forward progress. Let the leader's starting position be $P_{\text{start}}$, goal position be $P_{\text{goal}}$, and position at trial $k$ end be $P_{\text{end}}^k$. \textit{MP} is calculated as:
\begin{align*}
\text{\textit{MP}} = \frac{1}{N_{\text{trials}}} \sum_{k=1}^{N_{\text{trials}}} \Bigg[&\left(1 - \frac{\|P_{\text{end}}^k - P_{\text{goal}}\|_2}{\|P_{\text{start}} - P_{\text{goal}}\|_2}\right)\times 100\Bigg].
\end{align*}

\subsubsection{Flock Radius (\textit{FR})}
The maximum flock extent, defined as distance from swarm center to farthest UAV, averaged over the episode and trials. Compact formation ensures agents remain within perceptual range. Larger \textit{FR} indicates fragmentation risks. For $N$ UAVs, let $\mathbf{p}^{i,k}_t$ be position of UAV $i$ at time $t$ in trial $k$ and $\mathbf{p}^{c,k}_t = \frac{1}{N}\sum_{j=1}^{N} \mathbf{p}^{j,k}_t$ be the center. Let $T_k$ denote episode length of trial $k$. \textit{FR} is calculated by
\begin{align*}
\text{\textit{FR}} = \frac{1}{N_{\text{trials}}} \sum_{k=1}^{N_{\text{trials}}} \Bigg[&\frac{1}{T_k} \sum_{t=1}^{T_k} \max_{i=1,\dots,N} \|\mathbf{p}^{i,k}_t - \mathbf{p}^{c,k}_t\|_2\Bigg].
\end{align*}

\subsubsection{Minimum Separation (\textit{MS})}
A safety metric evaluating collision avoidance between agents, defined as minimum distance between the closest UAV pair, averaged over episode and trials. Adequate separation is fundamental to safe collective navigation.
\begin{align*}
\text{\textit{MS}} = \frac{1}{N_{\text{trials}}} \sum_{k=1}^{N_{\text{trials}}} \Bigg[&\frac{1}{T_k} \sum_{t=1}^{T_k} \min_{i \neq j} \|\mathbf{p}^{i,k}_t - \mathbf{p}^{j,k}_t\|_2\Bigg].
\end{align*}

\subsubsection{Alignment (\textit{AL})}
A metric measuring directional consistency, averaged over episode and trials. High \textit{AL} indicates coherent swarm movement. It is the average cosine similarity between each UAV's velocity $\mathbf{v}^{i,k}_t$ and swarm average velocity $\mathbf{v}^{\text{avg},k}_t = \frac{1}{N}\sum_{j=1}^{N} \mathbf{v}^{j,k}_t$. Values near $1$ indicate well-aligned formation.
\begin{align*}
\text{\textit{AL}} = \frac{1}{N_{\text{trials}}} \sum_{k=1}^{N_{\text{trials}}} \Bigg[&\frac{1}{T_k} \sum_{t=1}^{T_k} \frac{1}{N} \sum_{i=1}^{N} \frac{\mathbf{v}^{i,k}_t \cdot \mathbf{v}^{\text{avg},k}_t}{\|\mathbf{v}^{i,k}_t\|_2 \|\mathbf{v}^{\text{avg},k}_t\|_2}\Bigg].
\end{align*}

\subsubsection{Minimum Distance to Obstacles (\textit{MDO})}
A safety metric evaluating obstacle avoidance, defined as minimum distance between any UAV and any obstacle detection point, averaged over episode and trials. This measures safety margins maintained during navigation. Let $\mathbf{o}^{m,k}_t$ denote position of the $m$-th LiDAR detection point among $M$ total points:
\begin{align*}
\text{\textit{MDO}} = \frac{1}{N_{\text{trials}}} \sum_{k=1}^{N_{\text{trials}}} \Bigg[&\frac{1}{T_k} \sum_{t=1}^{T_k} \min_{\substack{i=1,\dots,N \\ m=1,\dots,M}} \|\mathbf{p}^{i,k}_t - \mathbf{o}^{m,k}_t\|_2\Bigg].
\end{align*}

\begin{table*}[t!]
    \centering
    \caption{Performance comparison with baseline methods}
    \label{tab:comparison}
    \resizebox{0.9\textwidth}{!}{%
    \begin{tabular}{@{}ll*{5}{C{2.2cm}}@{}}
    \toprule
    \multirow{5}{*}{\textbf{Metric}} & \multirow{5}{*}{\textbf{Method}} & \multirow{5}{*}{\textbf{No obstacles}} & \multicolumn{2}{c}{\textbf{Training environment}} & \multicolumn{2}{c}{\textbf{Test environment}} \\
    \cmidrule(lr){4-5} \cmidrule(lr){6-7}
    & & & \multicolumn{2}{c}{\textbf{Cluttered pillars}} & \textbf{Simplified forest} & \textbf{Forest} \\
    \cmidrule(lr){4-5} \cmidrule(lr){6-6} \cmidrule(lr){7-7}
    & & & \multicolumn{4}{c}{\textbf{Min. obstacle gap}} \\
    \cmidrule(lr){4-7}
    & & & \textbf{10m} & \textbf{5m} & \textbf{5m} & \textbf{5m} \\
    \midrule
    
    \multirow{4}{*}{\textbf{\textit{SR} (\%)}} 
    & PACNav~\cite{ahmad2022pacnav}     & 100.0$\pm$0.0 & 96.0$\pm$0.0 & 82.0$\pm$0.0 & 38.0$\pm$0.0 & 31.0$\pm$0.0 \\
    & VPF~\cite{wang2023collective}     & 100.0$\pm$0.0 & 80.0$\pm$0.0 & 27.0$\pm$0.0 & 6.0$\pm$0.0 & 5.0$\pm$0.0 \\
    & DAgger~\cite{wan2024distributed}  & 100.0$\pm$0.0 & 93.0$\pm$0.0 & 69.0$\pm$0.0 & 34.0$\pm$0.0 & 22.0$\pm$0.0 \\
    & Proposed                 & \textbf{100.0$\pm$0.0} & \textbf{99.0$\pm$0.0} & \textbf{97.0$\pm$0.0} & \textbf{97.0$\pm$0.0} & \textbf{72.0$\pm$0.0} \\
    \midrule
    
    \multirow{4}{*}{\textbf{\textit{MP} (\%)}} 
    & PACNav~\cite{ahmad2022pacnav}     & 100.0$\pm$0.0 & 98.6$\pm$8.8 & 93.0$\pm$18.6 & 72.2$\pm$26.6 & 67.6$\pm$28.2 \\
    & VPF~\cite{wang2023collective}     & 100.0$\pm$0.0 & 92.6$\pm$17.5 & 62.0$\pm$32.3 & 47.0$\pm$24.4 & 39.7$\pm$22.7 \\
    & DAgger~\cite{wan2024distributed}  & 100.0$\pm$0.0 & 97.1$\pm$11.4 & 85.6$\pm$25.4 & 74.5$\pm$25.9 & 69.1$\pm$24.0 \\
    & Proposed                 & \textbf{100.0$\pm$0.0} & \textbf{99.3$\pm$7.0} & \textbf{99.5$\pm$3.3} & \textbf{98.2$\pm$10.4} & \textbf{88.2$\pm$23.6} \\
    \midrule
    
    \multirow{4}{*}{\textbf{\textit{FR} (m)}} 
    & PACNav~\cite{ahmad2022pacnav}     & 1.80$\pm$0.14 & 1.83$\pm$0.12 & 1.85$\pm$0.10 & 1.85$\pm$0.10 & 1.82$\pm$0.10 \\
    & VPF~\cite{wang2023collective}     & 2.20$\pm$0.44 & 2.23$\pm$0.55 & 2.54$\pm$1.57 & 2.44$\pm$1.44 & 2.31$\pm$0.79 \\
    & DAgger~\cite{wan2024distributed}  & 2.96$\pm$0.44 & 2.85$\pm$0.44 & 2.65$\pm$0.37 & 2.59$\pm$0.56 & 2.59$\pm$0.46 \\
    & Proposed                 & \textbf{1.61$\pm$0.13} & \textbf{1.56$\pm$0.12} & \textbf{1.47$\pm$0.09} & \textbf{1.40$\pm$0.06} & \textbf{1.39$\pm$0.15} \\
    \midrule
    
    \multirow{4}{*}{\textbf{\textit{MS} (m)}} 
    & PACNav~\cite{ahmad2022pacnav}     & 1.23$\pm$0.07 & 1.25$\pm$0.06 & 1.26$\pm$0.06 & 1.27$\pm$0.04 & 1.25$\pm$0.05 \\
    & VPF~\cite{wang2023collective}     & 1.49$\pm$0.03 & 1.48$\pm$0.04 & 1.48$\pm$0.05 & 1.46$\pm$0.05 & 1.47$\pm$0.06 \\
    & DAgger~\cite{wan2024distributed}  & 1.54$\pm$0.11 & 1.53$\pm$0.11 & 1.51$\pm$0.11 & \textbf{1.49$\pm$0.13} & \textbf{1.49$\pm$0.13} \\
    & Proposed                 & \textbf{1.55$\pm$0.02} & \textbf{1.53$\pm$0.02} & \textbf{1.51$\pm$0.02} & 1.47$\pm$0.03 & 1.46$\pm$0.04 \\
    \midrule
    
    \multirow{4}{*}{\textbf{\textit{AL}}} 
    & PACNav~\cite{ahmad2022pacnav}     & 0.87$\pm$0.08 & 0.83$\pm$0.08 & 0.79$\pm$0.07 & 0.73$\pm$0.05 & 0.75$\pm$0.06 \\
    & VPF~\cite{wang2023collective}     & 0.92$\pm$0.03 & 0.92$\pm$0.04 & 0.89$\pm$0.06 & 0.89$\pm$0.05 & 0.87$\pm$0.09 \\
    & DAgger~\cite{wan2024distributed}  & 0.79$\pm$0.02 & 0.77$\pm$0.03 & 0.74$\pm$0.04 & 0.67$\pm$0.04 & 0.68$\pm$0.05 \\
    & Proposed                 & \textbf{0.94$\pm$0.01} & \textbf{0.94$\pm$0.01} & \textbf{0.93$\pm$0.02} & \textbf{0.90$\pm$0.02} & \textbf{0.88$\pm$0.11} \\
    \midrule

    \multirow{4}{*}{\textbf{\textit{MDO} (m)}} 
    & PACNav~\cite{ahmad2022pacnav}     & - & 2.18$\pm$0.33 & \textbf{1.89$\pm$0.40} & 0.82$\pm$0.45 & \textbf{0.82$\pm$0.32} \\
    & VPF~\cite{wang2023collective}     & - & 2.12$\pm$0.41 & 1.81$\pm$0.57 & 1.27$\pm$0.45 & 0.73$\pm$0.45 \\
    & DAgger~\cite{wan2024distributed}  & - & 1.89$\pm$0.40 & 1.52$\pm$0.54 & 0.93$\pm$0.58 & 0.76$\pm$0.41 \\
    & Proposed                 & - & \textbf{2.27$\pm$0.15} & 1.88$\pm$0.16 & \textbf{1.56$\pm$0.09} & 0.79$\pm$0.32 \\
    \bottomrule
    
    \end{tabular}%
    }
    \vspace{-12pt}
\end{table*}

\subsection{Performance Comparison}

To demonstrate the superiority and practical applicability of the proposed method, we conducted comparative evaluations against baseline methods satisfying two critical criteria: compatibility with real-world deployment on physical UAV platforms with actual sensor constraints and demonstrated effectiveness in communication-free collective navigation scenarios.

\subsubsection{Baseline Methods}
Our baseline selection comprises three representative approaches meeting these criteria. The first two are heuristic-based methods validated in real-world experiments. PACNav~\cite{ahmad2022pacnav} is a bio-inspired distributed control method enabling collective navigation without GNSS or inter-UAV communication by introducing two metrics: path persistence (measuring how consistently a UAV flies straight) and path similarity (measuring directional consistency among neighbors). Uninformed UAVs identify reliable leaders by selecting neighbors exhibiting high path persistence and similarity, enabling emergent collective navigation toward unknown destinations. The method incorporates reactive collision avoidance for environmental obstacles and inter-UAV conflicts, validated with four UAVs in natural forests.

The second baseline employs a bio-inspired VPF~\cite{wang2023collective}, which mimics avian visual perception for communication-free flocking. Instead of exchanging state information, each UAV uses VPF to observe neighbors and obstacles, distinguishing between repulsion zones (objects appearing large) and attraction zones (objects appearing small) based on perceived size. This enables the three fundamental flocking behaviors—separation, attraction, and velocity alignment—without inter-UAV communication. The approach incorporates an implicit heterogeneous flocking framework, where a small minority of informed UAVs guide the uninformed majority through emergent collective behavior, with uninformed agents following the swarm's overall motion without explicitly identifying leaders. The method has been validated with six UAVs in indoor environments with obstacles, where a motion capture system was installed.

For the learning-based baseline, the DAgger approach~\cite{wan2024distributed} trains an end-to-end visuomotor controller by imitating an expert policy combining modified Reynolds' flocking rules with APF-based obstacle avoidance. The expert policy, which has access to ground-truth state information in simulation, generates optimal control commands that the student policy learns to replicate using only onboard camera images and IMU data. Through centralized training and decentralized execution, the learned policy achieves vision-only cooperative flight without GNSS or inter-UAV communication. While validated exclusively in high-fidelity Gazebo simulations with complex scenarios (pillar forests, narrow gaps), the method demonstrates superior learning efficiency and performance compared with other learning algorithms (MADDPG~\cite{yan2023collision}, SAC~\cite{bai2023learning}, and MAGAIL~\cite{fang2022autonomous}). For our comparison, we directly implemented the expert policy itself rather than retraining the imitation learning policy. This design ensures fair comparison by eliminating approximation error from imitation learning, evaluating the fundamental control strategy rather than policy distillation quality, providing an upper bound on DAgger method's performance.

\subsubsection{Fair Comparison Setup}
For fair comparison, we unified all methods to use identical LiDAR-based perception (same FOV, range, occlusion handling) while preserving each method's control strategy. Parameters were independently tuned to achieve similar flocking behavior in obstacle-free environments, ensuring equivalent baseline capability before testing in complex scenarios.

\begin{table*}[t!]
\centering
\caption{Scalability analysis of the proposed DRL policy with varying numbers of follower UAVs}
\label{tab:scalability}
\resizebox{0.9\textwidth}{!}{%
\begin{tabular}{@{}ll*{5}{C{2.0cm}}@{}}
\toprule
\multirow{2}{*}{\textbf{Metric}} & \multirow{2}{*}{\textbf{Min. obstacle gap}} & \multicolumn{5}{c}{\textbf{Number of followers}} \\
\cmidrule(lr){3-7}
& & \textbf{2} & \textbf{4} & \textbf{6} & \textbf{8} & \textbf{10} \\
\midrule
\multirow{3}{*}{\textbf{\textit{SR} (\%)}} 
 & No obstacles & 100.0$\pm$0.0 & 100.0$\pm$0.0 & 100.0$\pm$0.0 & 98.0$\pm$0.0 & 93.0$\pm$0.0 \\
 & 10m          & 100.0$\pm$0.0 & 99.0$\pm$0.0 & 96.0$\pm$0.0 & 84.0$\pm$0.0 & 70.0$\pm$0.0 \\
 & 5m           & 93.0$\pm$0.0 & 97.0$\pm$0.0 & 88.0$\pm$0.0 & 56.0$\pm$0.0 & 46.0$\pm$0.0 \\
\midrule

\multirow{3}{*}{\textbf{\textit{MP} (\%)}} 
 & No obstacles & 100.0$\pm$0.0 & 100.0$\pm$0.0 & 100.0$\pm$0.0 & 99.2$\pm$7.5 & 95.3$\pm$18.2 \\
 & 10m          & 100.0$\pm$0.0 & 99.3$\pm$7.0 & 97.7$\pm$12.8 & 91.7$\pm$21.9 & 88.5$\pm$23.4 \\
 & 5m           & 98.0$\pm$9.7 & 99.5$\pm$3.3 & 96.8$\pm$10.2 & 79.6$\pm$22.9 & 70.3$\pm$27.6 \\
\midrule

\multirow{3}{*}{\textbf{\textit{FR} (m)}} 
 & No obstacles & 1.13$\pm$0.18 & 1.61$\pm$0.13 & 1.93$\pm$0.06 & 2.15$\pm$0.06 & 2.32$\pm$0.05 \\
 & 10m          & 1.08$\pm$0.15 & 1.56$\pm$0.12 & 1.88$\pm$0.06 & 2.14$\pm$0.06 & 2.33$\pm$0.06 \\
 & 5m           & 1.04$\pm$0.13 & 1.47$\pm$0.09 & 1.85$\pm$0.05 & 2.12$\pm$0.05 & 2.32$\pm$0.07 \\
\midrule

\multirow{3}{*}{\textbf{\textit{MS} (m)}} 
 & No obstacles & 1.60$\pm$0.01 & 1.55$\pm$0.02 & 1.47$\pm$0.03 & 1.39$\pm$0.02 & 1.34$\pm$0.02 \\
 & 10m          & 1.60$\pm$0.01 & 1.53$\pm$0.02 & 1.46$\pm$0.03 & 1.38$\pm$0.02 & 1.34$\pm$0.02 \\
 & 5m           & 1.60$\pm$0.01 & 1.51$\pm$0.02 & 1.44$\pm$0.03 & 1.37$\pm$0.02 & 1.32$\pm$0.03 \\
\midrule

\multirow{3}{*}{\textbf{\textit{AL}}} 
 & No obstacles & 0.97$\pm$0.02 & 0.94$\pm$0.01 & 0.89$\pm$0.02 & 0.85$\pm$0.02 & 0.84$\pm$0.04 \\
 & 10m          & 0.97$\pm$0.02 & 0.94$\pm$0.01 & 0.89$\pm$0.02 & 0.85$\pm$0.02 & 0.84$\pm$0.03 \\
 & 5m           & 0.97$\pm$0.02 & 0.93$\pm$0.02 & 0.88$\pm$0.02 & 0.84$\pm$0.03 & 0.83$\pm$0.04 \\
\midrule

\multirow{3}{*}{\textbf{\textit{MDO} (m)}} 
 & No obstacles & - & - & - & - & - \\
 & 10m          & 2.45$\pm$0.13 & 2.27$\pm$0.15 & 2.17$\pm$0.18 & 2.00$\pm$0.24 & 1.68$\pm$0.29 \\
 & 5m           & 2.12$\pm$0.21 & 1.88$\pm$0.16 & 1.65$\pm$0.29 & 1.49$\pm$0.37 & 1.35$\pm$0.45 \\
\bottomrule

\end{tabular}%
}
\vspace{-12pt}
\end{table*}

\subsubsection{Evaluation Environments}
Comparative evaluation was conducted with five UAVs (one leader and four followers) in various environments to verify generalization performance and robustness. First, evaluations were performed in the training environment with three obstacle density levels: none, sparse with minimum 10m gap, and dense with minimum 5m gap. Two new test environments were introduced: a simplified forest with irregularly distributed trees and rocks, and a more realistic complex forest featuring trees and rocks with detailed trunks and foliage as shown in Fig.~\ref{fig:simulation}(b) and (c).

\begin{figure}[t!]
    \centering
    \includegraphics[width=\columnwidth]{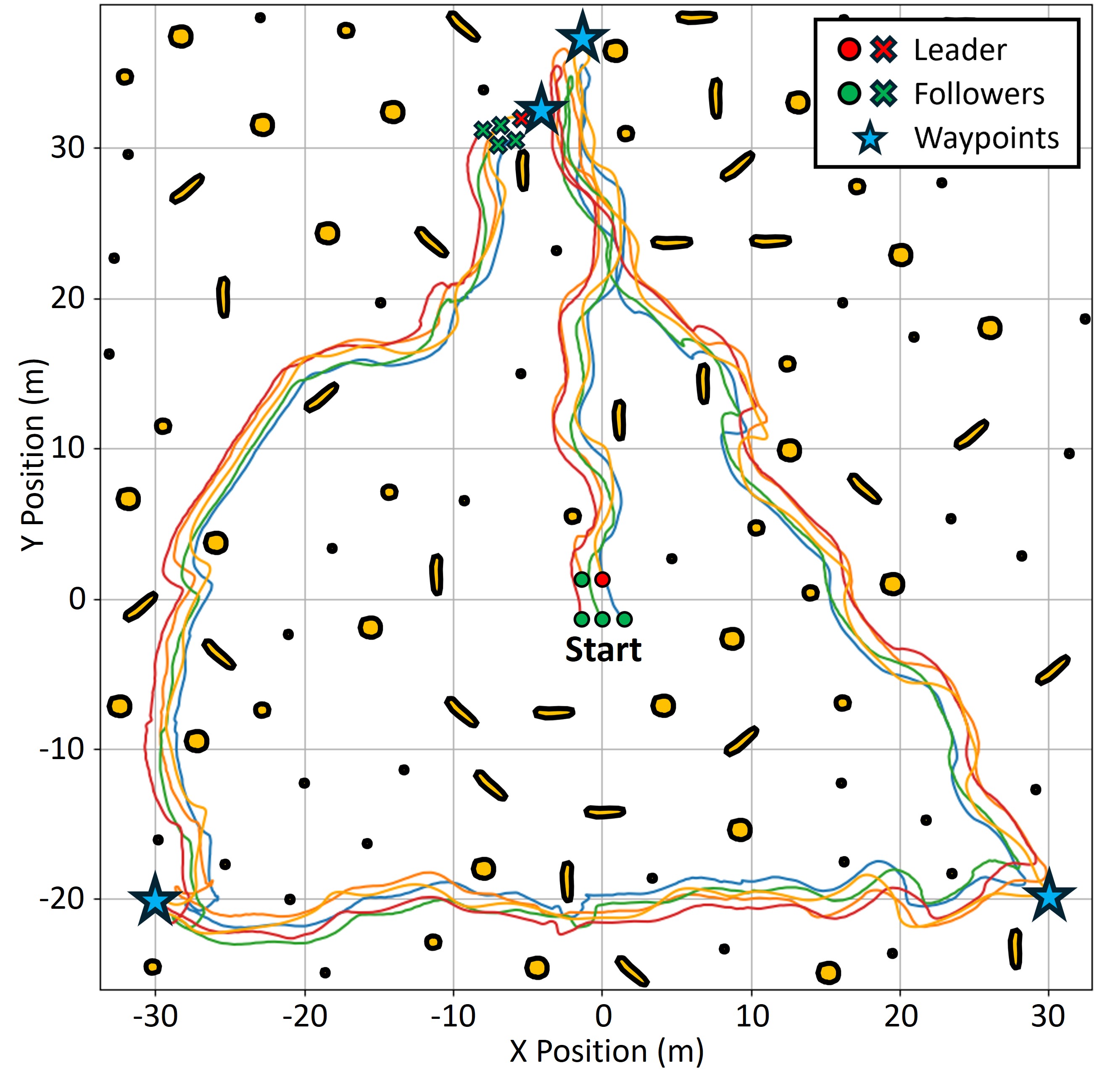}
    \vspace{-20pt}
    \caption{Example trajectory of the proposed policy in the forest test environment with multiple waypoints. The swarm of five UAVs successfully navigates through dense obstacles by following the leader while maintaining cohesion. The leader sequentially visits four waypoints to reach the goal.}
    \label{fig:forest_trajectory}
    \vspace{-12pt}
\end{figure}

\subsubsection{Results and Analysis}
As shown in Table~\ref{tab:comparison}, the proposed DRL method consistently outperforms all baselines, with performance gaps widening in complex environments. All methods succeed without obstacles, but degrade sharply in cluttered scenarios (5--31\% success in realistic forests), whereas our policy maintains 72\% success and 88.2\% mission progress. The proposed method achieves the most compact flock (smallest radius) and organized formation (highest alignment) while maintaining safe separation and obstacle distances, effectively balancing task completion, cohesion, and collision avoidance. 

Beyond quantitative evaluation with single goal points, Fig.~\ref{fig:forest_trajectory} demonstrates the proposed policy's versatility in handling sequential waypoint navigation. The swarm successfully follows the leader through four waypoints in the forest test environment, maintaining cohesion and obstacle avoidance. This capability naturally emerges from the learned flocking behavior and obstacle avoidance, as followers implicitly follow the leader regardless of waypoint number or configuration.

\subsection{Scalability Analysis}

To evaluate the scalability of the proposed DRL policy, a single model trained with five UAVs (one leader and four followers) was tested by varying the number of followers from two to ten. Table~\ref{tab:scalability} illustrates performance variation with respect to follower count and environmental complexity.

Success rate and mission progress decreased with increasing swarm size and environmental complexity, with significant degradation when eight or more followers navigated narrow gaps. This decline stems from a geometric constraint: the policy maintains similar altitudes for reliable neighbor detection within LiDAR's limited vertical FOV ($-7^\circ$ to $52^\circ$), causing horizontal rather than vertical formation expansion. With 10 followers, the flock radius reaches 2.32m (total diameter ~4.6m), making navigation through 5m gaps physically challenging. This tight margin between swarm size and obstacle spacing causes high failure rates in large-scale swarms.

Flock radius increased with swarm size regardless of environment, while minimum separation and alignment decreased. Reduced separation implies higher collision risk, and decreased alignment reveals a key limitation of decentralized control: difficulty achieving directional consensus in large swarms without communication. The policy demonstrates moderate scalability but degrades due to geometric constraints, suggesting that robust large-scale application requires policies that better utilize three-dimensional space.

\subsection{Ablation Studies} 
\label{sec:ablation}

\subsubsection{Effect of Reward Components}

We validated each reward term by systematically removing components and evaluating six configurations: (1) without flocking reward, (2) without obstacle avoidance, (3) without stable flight, (4) without neighbor perception, (5) uniform weights, and (6) proposed method with tuned weights. Each was trained for 500 million timesteps and tested across three obstacle densities.

\begin{table*}[t!]
\centering
\caption{Ablation study of reward components across environments with varying obstacle densities}
\label{tab:reward_ablation}
\resizebox{\textwidth}{!}{%
\begin{tabular}{@{}ll*{6}{C{2cm}}@{}}
\toprule
\multirow{2}{*}{\textbf{Metric}} & \multirow{2}{*}{\textbf{Min. obstacle gap}} & \multicolumn{6}{c}{\textbf{Configuration}} \\
\cmidrule(lr){3-8}
& & \textbf{w/o $r_{\text{flocking}}$} & \textbf{w/o $r_{\text{obstacle}}$} & \textbf{w/o $r_{\text{stable}}$} & \textbf{w/o $r_{\text{perception}}$} & \textbf{Uniform} & \textbf{Proposed} \\
\midrule
\multirow{3}{*}{\textbf{\textit{SR} (\%)}} 
 & No obstacles & 0.0$\pm$0.0 & 96.0$\pm$0.0 & 11.0$\pm$0.0 & 99.0$\pm$0.0 & 82.0$\pm$0.0 & \textbf{100.0$\pm$0.0} \\
 & 10m          & 0.0$\pm$0.0 & 96.0$\pm$0.0 & 9.0$\pm$0.0 & 96.0$\pm$0.0 & 74.0$\pm$0.0 & \textbf{99.0$\pm$0.0} \\
 & 5m           & 0.0$\pm$0.0 & 86.0$\pm$0.0 & 3.0$\pm$0.0 & 67.0$\pm$0.0 & 54.0$\pm$0.0 & \textbf{97.0$\pm$0.0} \\
\midrule

\multirow{3}{*}{\textbf{\textit{MS} (\%)}} 
 & No obstacles & - & 97.8$\pm$11.8 & 28.0$\pm$39.8 & 99.2$\pm$8.5 & 92.6$\pm$20.9 & \textbf{100.0$\pm$0.0} \\
 & 10m          & - & 99.2$\pm$5.8 & 32.5$\pm$39.0 & 98.9$\pm$6.4 & 92.0$\pm$22.6 & \textbf{99.3$\pm$7.0} \\
 & 5m           & - & 94.2$\pm$16.3 & 35.7$\pm$29.3 & 81.3$\pm$30.3 & 80.0$\pm$32.5 & \textbf{99.5$\pm$3.3} \\
\midrule

\multirow{3}{*}{\textbf{\textit{FR} (m)}} 
 & No obstacles & - & 2.02$\pm$0.11 & 3.60$\pm$0.87 & 2.29$\pm$0.15 & 2.85$\pm$0.26 & \textbf{1.61$\pm$0.13} \\
 & 10m          & - & 2.03$\pm$0.10 & 3.35$\pm$0.79 & 2.26$\pm$0.14 & 2.86$\pm$0.25 & \textbf{1.56$\pm$0.12} \\
 & 5m           & - & 2.00$\pm$0.09 & 2.94$\pm$0.78 & 2.23$\pm$0.14 & 2.83$\pm$0.29 & \textbf{1.47$\pm$0.09} \\
\midrule

\multirow{3}{*}{\textbf{\textit{MS} (m)}} 
 & No obstacles & - & \textbf{1.58$\pm$0.02} & 1.55$\pm$0.05 & 1.53$\pm$0.04 & 1.42$\pm$0.07 & 1.55$\pm$0.02 \\
 & 10m          & - & \textbf{1.58$\pm$0.01} & 1.55$\pm$0.04 & 1.52$\pm$0.03 & 1.40$\pm$0.06 & 1.53$\pm$0.02 \\
 & 5m           & - & \textbf{1.57$\pm$0.02} & 1.55$\pm$0.04 & 1.50$\pm$0.04 & 1.38$\pm$0.05 & 1.51$\pm$0.02 \\
\midrule

\multirow{3}{*}{\textbf{\textit{AL}}} 
 & No obstacles & - & 0.94$\pm$0.02 & 0.90$\pm$0.04 & 0.94$\pm$0.01 & \textbf{0.95$\pm$0.02} & 0.94$\pm$0.01 \\
 & 10m          & - & 0.94$\pm$0.01 & 0.90$\pm$0.04 & 0.94$\pm$0.01 & 0.94$\pm$0.02 & \textbf{0.94$\pm$0.01} \\
 & 5m           & - & \textbf{0.94$\pm$0.02} & 0.90$\pm$0.04 & 0.92$\pm$0.03 & 0.93$\pm$0.03 & 0.93$\pm$0.02 \\
\midrule

\multirow{3}{*}{\textbf{\textit{MDO} (m)}} 
 & No obstacles & - & - & - & - & - & - \\
 & 10m          & - & 2.25$\pm$0.18 & 2.22$\pm$0.17 & \textbf{2.41$\pm$0.15} & 2.36$\pm$0.14 & 2.27$\pm$0.15 \\
 & 5m           & - & 1.66$\pm$0.26 & 1.86$\pm$0.30 & 1.82$\pm$0.27 & 1.74$\pm$0.20 & \textbf{1.88$\pm$0.16} \\
\bottomrule

\end{tabular}%
}
\vspace{-12pt}
\end{table*}

As shown in Table~\ref{tab:reward_ablation}, removing flocking reward completely destroys collective navigation (0\% success), with followers hovering at initial positions while the leader moves away. Without cohesion incentives, followers cannot establish leader-following behavior, demonstrating this reward is fundamentally essential for implicit coordination in communication-free settings.

Removing obstacle avoidance shows minor degradation in simple environments (96\% success rate) but drops to 86\% in dense scenarios, indicating explicit guidance significantly improves robustness despite some learning through collision penalties. The stable flight reward proves critical—its removal causes catastrophic failure (11\%, 9\%, and 3\% success rates) as followers drift in altitude, losing neighbors outside LiDAR's limited FOV ($-7^\circ$ to $52^\circ$), with increased flock radius (2.94m to 3.60m) and erratic flight patterns. This highlights altitude consistency as prerequisite for effective coordination with limited FOV sensors.

The neighbor perception reward shows environment-dependent importance, with minimal impact in simple scenarios (99\% and 96\%) but significant drops to 67\% in dense environments, suggesting visibility mechanisms become critical as occlusion increases. Uniform weights achieve moderate performance (82\%, 74\%, and 54\%) but consistently underperform the proposed method (100\%, 99\%, and 97\%). These results validate each component's distinct role: flocking enables leader-following, obstacle avoidance ensures safety, stable flight maintains controllability, and perception enhances robustness, with tuned weights effectively balancing these competing objectives.

\subsubsection{Effect of Observable Neighbors}

\begin{figure}[t!]
    \centering
    \includegraphics[width=\columnwidth]{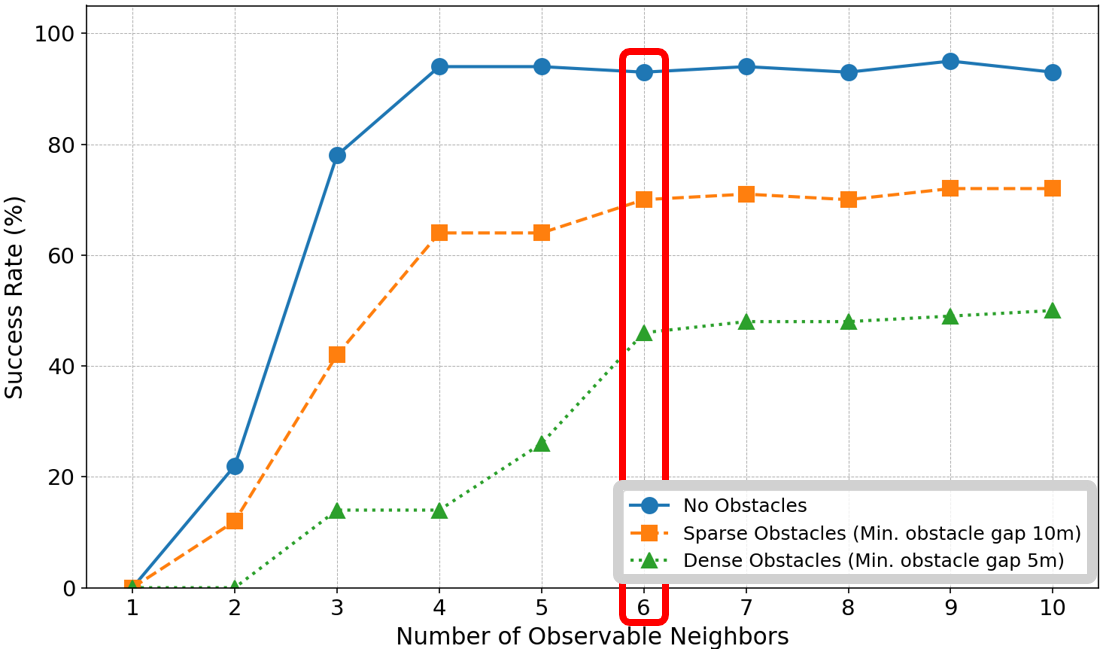}
    \vspace{-20pt}
    \caption{An effect of observable neighbors on success rate in collective navigation.}
    \label{fig:ablation_study}
    \vspace{-12pt}
\end{figure}

To determine the appropriate number of observable neighbors, we performed an ablation study using eleven UAVs (one leader and ten followers) across environments with varying complexity. The results, shown in Fig.~\ref{fig:ablation_study}, indicate a clear trend: while increasing the number of neighbors generally improves success rate, performance gains diminish after a certain point.

In the obstacle-free environment, success rate saturated at 94\% with only four neighbors, showing no further improvement with additional information. In more challenging environments with obstacles, a larger number of neighbors proved beneficial, though a noticeable slowdown in performance gains was observed. For instance, in the dense obstacles environment (5m spacing), increasing observable neighbors from five to six yielded a substantial 20\% increase in success rate (from 26\% to 46\%). In contrast, further increasing from six to ten provided only a marginal 4\% gain (from 46\% to 50\%), after which performance showed no significant improvement.

This analysis reveals that $|\mathcal{N}|=6$ is a critical threshold capturing the majority of performance benefits from situational awareness, particularly in complex scenarios. Beyond this number, marginal gains in success rate are outweighed by increased computational cost. Specifically, each additional neighbor contributes new data points, and the computational load of the object tracker, which scales quadratically ($O(n^2)$) with the number of points $n$, becomes a significant bottleneck on onboard embedded systems. Therefore, setting observable neighbors to six provides balance, ensuring high performance across various environments without imposing unnecessary computational burdens on the agents.

\begin{figure*}[t!]
    \centering
    \includegraphics[width=0.8\textwidth]{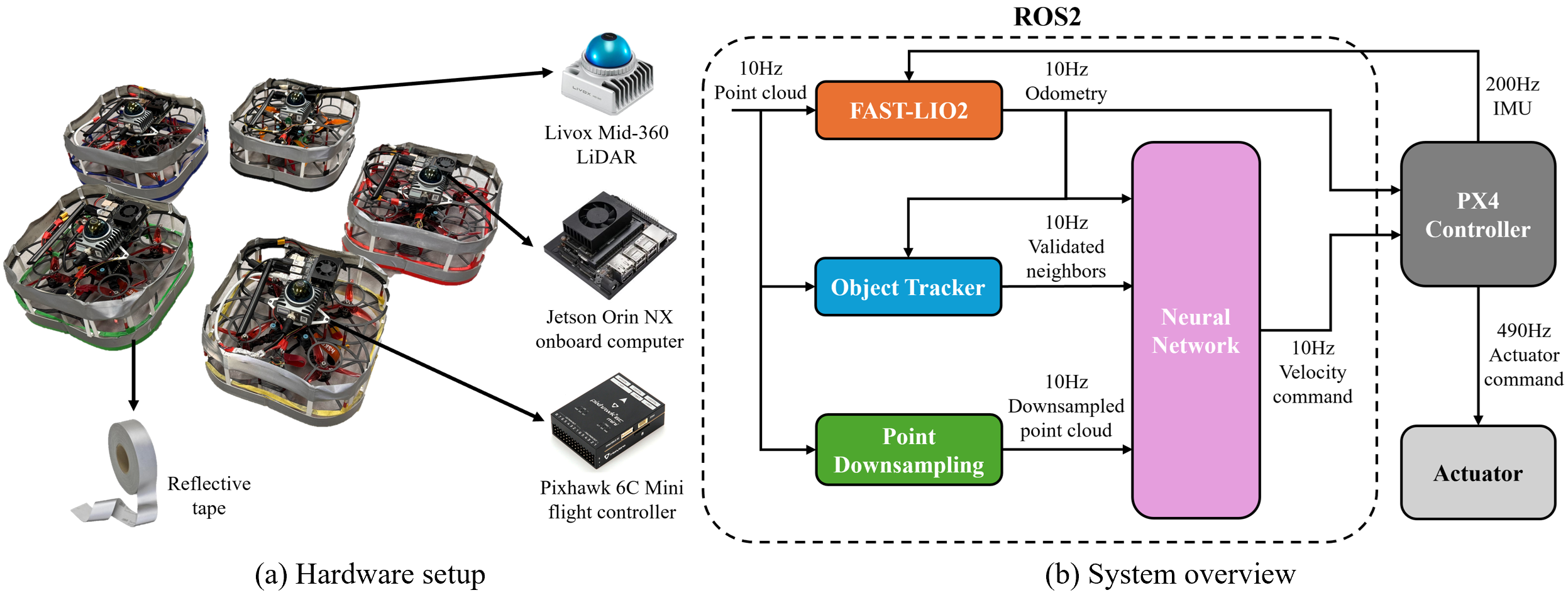}
    \vspace{-10pt}
    \caption{Hardware and system overview. (a) A custom quadrotor equipped with a LiDAR, onboard computer, and reflective tape. (b) The onboard software architecture, showing the flow from sensor data to actuator commands.}
    \label{fig:system_overview}
\end{figure*}

\section{Experiments}
\label{sec:experiments}

\begin{figure*}[t!]
    \centering
    \includegraphics[width=0.7\textwidth]{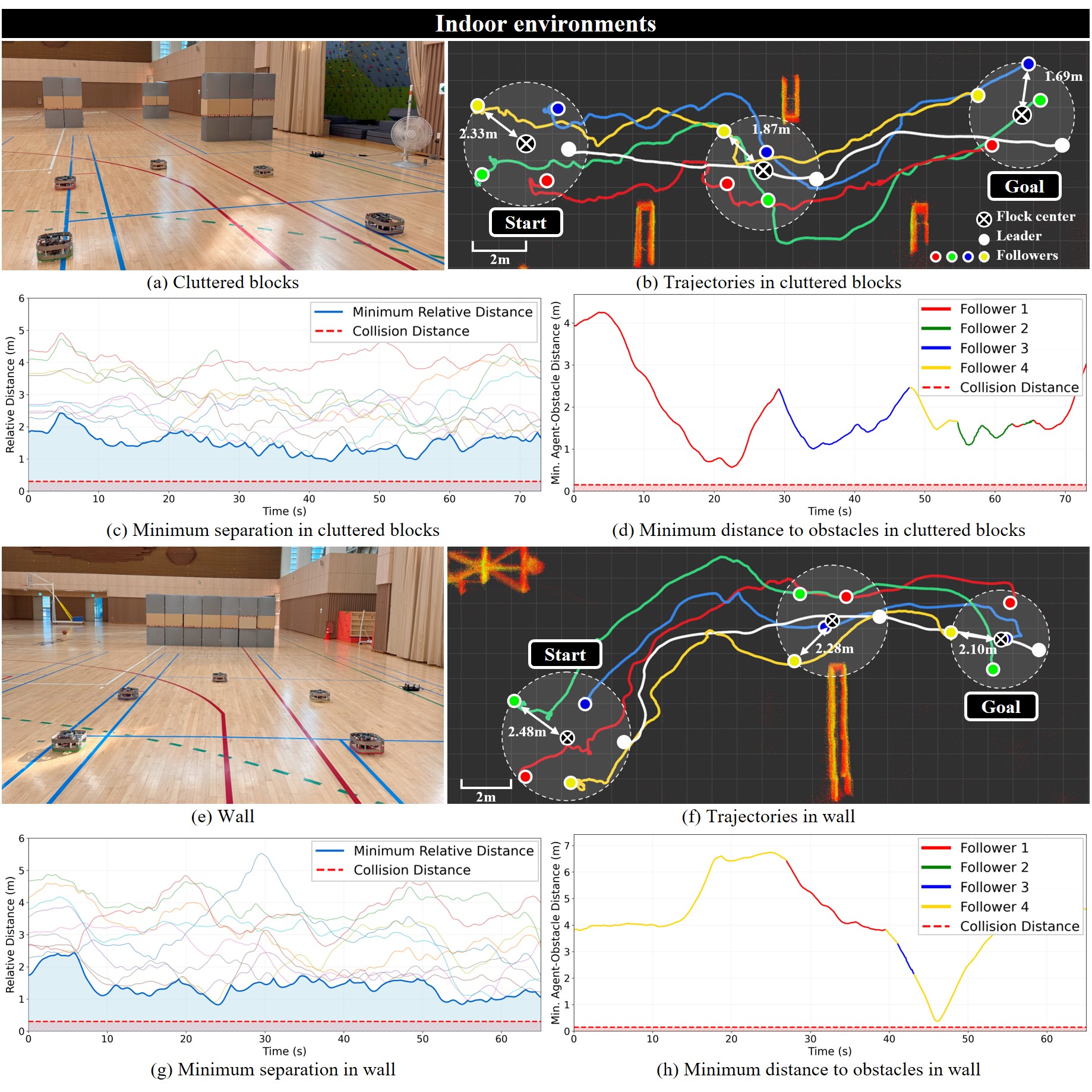}
    \vspace{-10pt}
    \caption{Real-world validation in indoor environments. (a), (e) Experimental setups for cluttered blocks and wall. (b), (f) Corresponding swarm trajectories visualizing the flock radius. (c), (g) Minimum separation and (d), (h) Minimum distance to obstacles over time, respectively.}
    \label{fig:indoor}
    \vspace{-12pt}
\end{figure*}

\begin{figure*}[t!]
    \centering
    \includegraphics[width=0.7\textwidth]{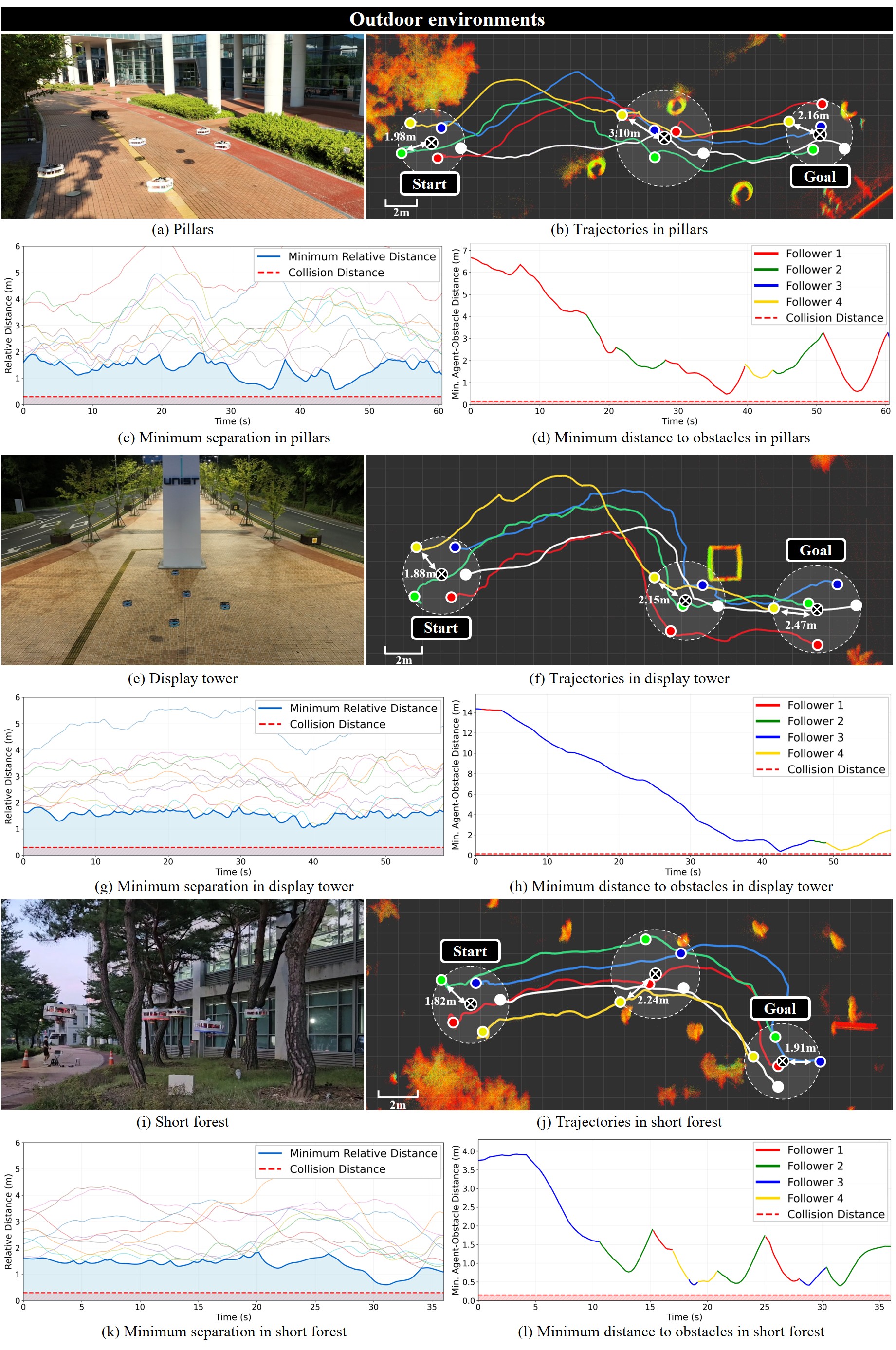}
    \vspace{-10pt}
    \caption{Real-world validation in outdoor environments. (a), (e), (i) Experimental setups for pillars, display tower, and short forest. (b), (f), (j) Corresponding swarm trajectories visualizing the flock radius. (c), (g), (k) Minimum separation and (d), (h), (l) Minimum distance to obstacles over time, respectively.}
    \label{fig:outdoor}
    \vspace{-12pt}
\end{figure*}

\subsection{System Overview} 

We used custom 250\,mm quadrotors equipped with a Livox Mid-360 LiDAR, an Nvidia Jetson Orin NX, and a Pixhawk 6C flight controller as shown in Fig.~\ref{fig:system_overview}. To ensure the UAV is detectable from any angle, reflective tape was affixed to all sides of its body. All computation, including FAST-LIO2~\cite{xu2022fast} for ego-pose estimation, an object tracker for neighbor detection and tracking, point downsampling, and neural network inference, was performed onboard.

\subsection{Performance of LiDAR-based detection} 

We evaluated the LiDAR-based detection performance under various conditions. The number of LiDAR returns decreases inversely with the square of the distance, and beyond approximately 10 m the low point density prevented reliable detection for our UAV size. In addition, the reflective tape’s intensity drops sharply at incident angles greater than about 45 degrees, leading to missed detections due to insufficient intensity. To address these limitations, reflective tape was affixed to all sides of the UAV. 

Detection rate is defined as the proportion of instances in which a UAV present in the scene is successfully detected by the LiDAR and precision represents the proportion of LiDAR detections that correspond to actual UAVs. Excluding cases where neighboring UAVs were outside the LiDAR’s FOV, our experiments demonstrated a detection rate of 100\% and a precision of 99.2\%, measured as the ratio of correctly detected instances across all time steps.

\subsection{Experimental Results}

To validate the sim-to-real transfer, we conducted experiments in controlled indoor and unstructured outdoor environments, with results in Fig.~\ref{fig:indoor}, Fig.~\ref{fig:outdoor}, and Table~\ref{tab:real_world_exp}. The results presented for each environment were obtained from a single trial, where the reported metrics represent the temporal mean and standard deviation over the entire episode length. The successful sim-to-real transfer of our policy was clearly demonstrated in diverse real-world environments. In indoor trials, the swarm successfully navigated cluttered obstacles while maintaining robust cohesion and separation. The low alignment scores observed were a direct result of confined space, which necessitated frequent maneuvering for collision avoidance. Both the minimum separation between agents and minimum distance to obstacles consistently remained above the collision distance threshold throughout all experiments, ensuring safe navigation without any collisions. Meanwhile, in outdoor scenarios with unpredictable variables such as wind and varying lighting conditions, the swarm demonstrated robust performance with these safety margins maintained across diverse obstacle configurations. The short forest scenario presented the most challenging obstacle avoidance conditions due to its dense vegetation, yet the policy successfully maintained collision-free navigation. The higher alignment scores in open areas suggest the policy adaptively transitions to a more organized and efficient formation when space allows. Collectively, these results validate that the single policy exhibits excellent generalization and robustness under complex and various real-world conditions.

\newcolumntype{C}[1]{>{\centering\arraybackslash}m{#1}}
\newcolumntype{L}[1]{>{\raggedright\arraybackslash}m{#1}} 

\begin{table*}[t!]
\centering
\caption{Real-world experiment results with five UAVs}
\label{tab:real_world_exp}
\resizebox{0.7\textwidth}{!}{%
\begin{tabular}{@{}L{1.2cm}*{5}{C{2cm}}@{}}
\toprule
\multirow{2}{*}{\textbf{Metric}} & \multicolumn{2}{c}{\textbf{Indoor environments}} & \multicolumn{3}{c}{\textbf{Outdoor environments}} \\
\cmidrule(lr){2-3} \cmidrule(lr){4-6}
& \textbf{Cluttered blocks} & \textbf{Wall} & \textbf{Pillars} & \textbf{Display tower} & \textbf{Short forest} \\
\midrule

\textbf{\textit{FR} (m)} & 2.31$\pm$0.28 & 2.54$\pm$0.28 & 2.86$\pm$0.53 & 2.60$\pm$0.27 & 2.30$\pm$0.30 \\

\textbf{\textit{MS} (m)} & 1.50$\pm$0.28 & 1.39$\pm$0.36 & 1.34$\pm$0.34 & 1.55$\pm$0.16 & 1.36$\pm$0.27 \\

\textbf{\textit{AL}} & 0.48$\pm$0.40 & 0.52$\pm$0.42 & 0.72$\pm$0.34 & 0.81$\pm$0.21 & 0.74$\pm$0.30 \\

\textbf{\textit{MDO} (m)} & 1.88$\pm$0.92 & 4.25$\pm$1.45 & 2.82$\pm$1.83 & 5.82$\pm$4.69 & 1.57$\pm$1.13 \\

\bottomrule
\end{tabular}%
}
\vspace{-12pt}
\end{table*}


\section{Conclusions}

This paper introduced a fully communication-free system for collective UAV swarm navigation using a single LiDAR and a DRL policy. Our implicit leader-follower framework, enabled by a robust perception pipeline and a learned controller, allows a swarm to perform complex navigation in communication-denied environments. We rigorously validated our approach through extensive simulations and challenging real-world experiments. The DRL-based policy outperformed existing methods in simulation and demonstrated successful sim-to-real transfer, with a five-UAV swarm navigating diverse indoor and outdoor environments. This study validated the practicality and robustness of using DRL for communication-free collective navigation. 

Future work will focus on enhancing scalability to larger swarms and exploring more complex collective behaviors like adaptive role-switching.

\section*{Author Contributions}
Myong-Yol Choi: Conceptualization, project administration, investigation, methodology, software, validation, data curation, formal analysis, visualization, and writing—original draft.
Hankyoul Ko: Investigation, software, validation, data curation, formal analysis, and visualization.
Hanse Cho: Investigation, data curation, and formal analysis.
Changseung Kim, Seunghwan Kim, and Jaemin Seo: Investigation.
Hyondong Oh: Supervision, funding acquisition, resources, and writing—review and editing.


\bibliographystyle{IEEEtran}
\bibliography{references.bib}

\begin{IEEEbiography}[{\includegraphics
[width=1in,height=1.25in,clip,
keepaspectratio]{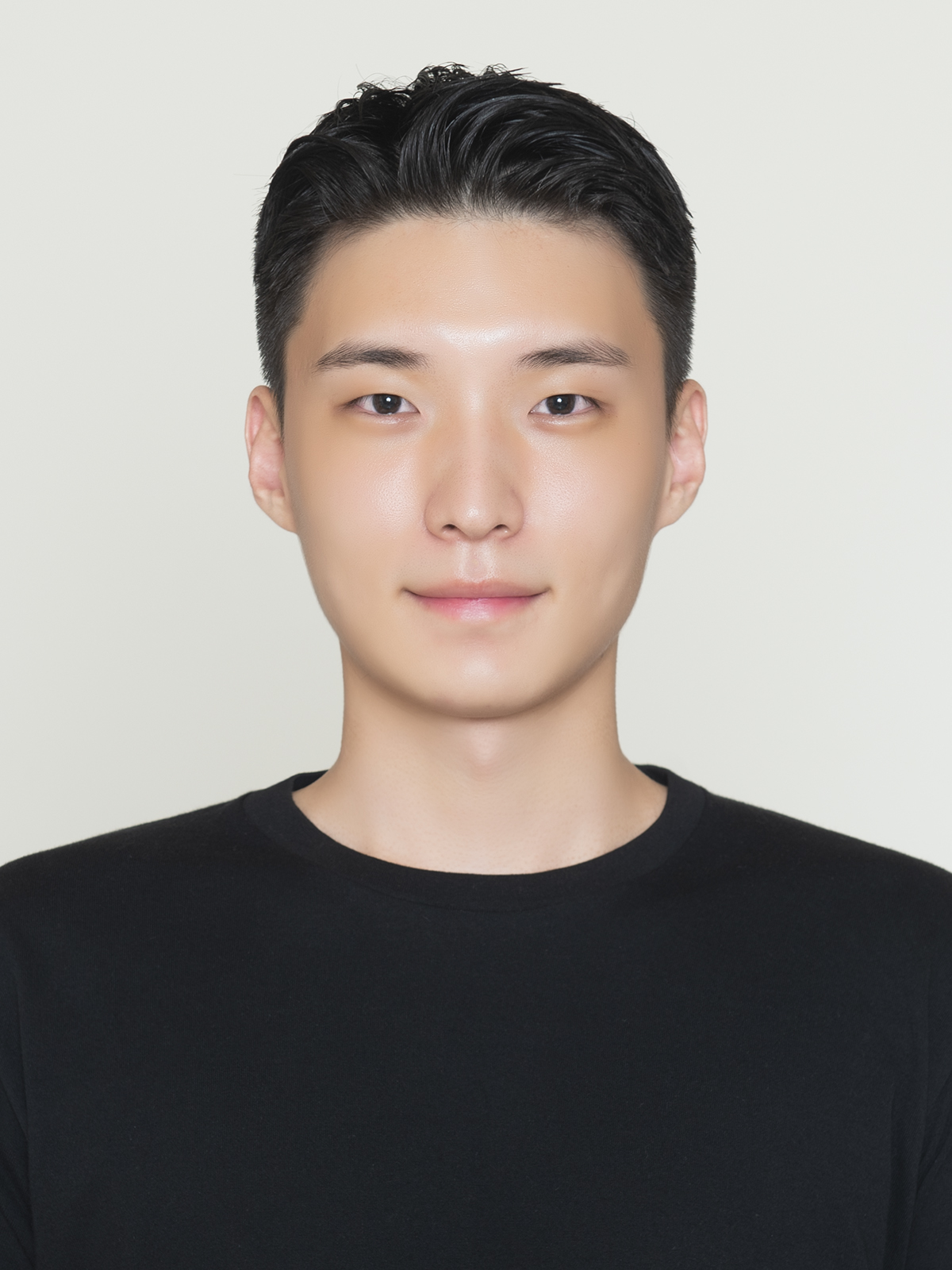}}]
{Myong-Yol Choi} is currently a Ph.D. candidate at the Department of Mechanical Engineering, Ulsan National Institute of Science and Technology, Ulsan, Republic of Korea. He received a B.S. degree in Mechanical and Aerospace Engineering from Ulsan National Institute of Science and Technology, Ulsan, Republic of Korea, in 2022. His research focuses on learning-based collective behaviors for autonomous swarms of unmanned vehicles.
\end{IEEEbiography}

\begin{IEEEbiography}[{\includegraphics
[width=1in,height=1.25in,clip,
keepaspectratio]{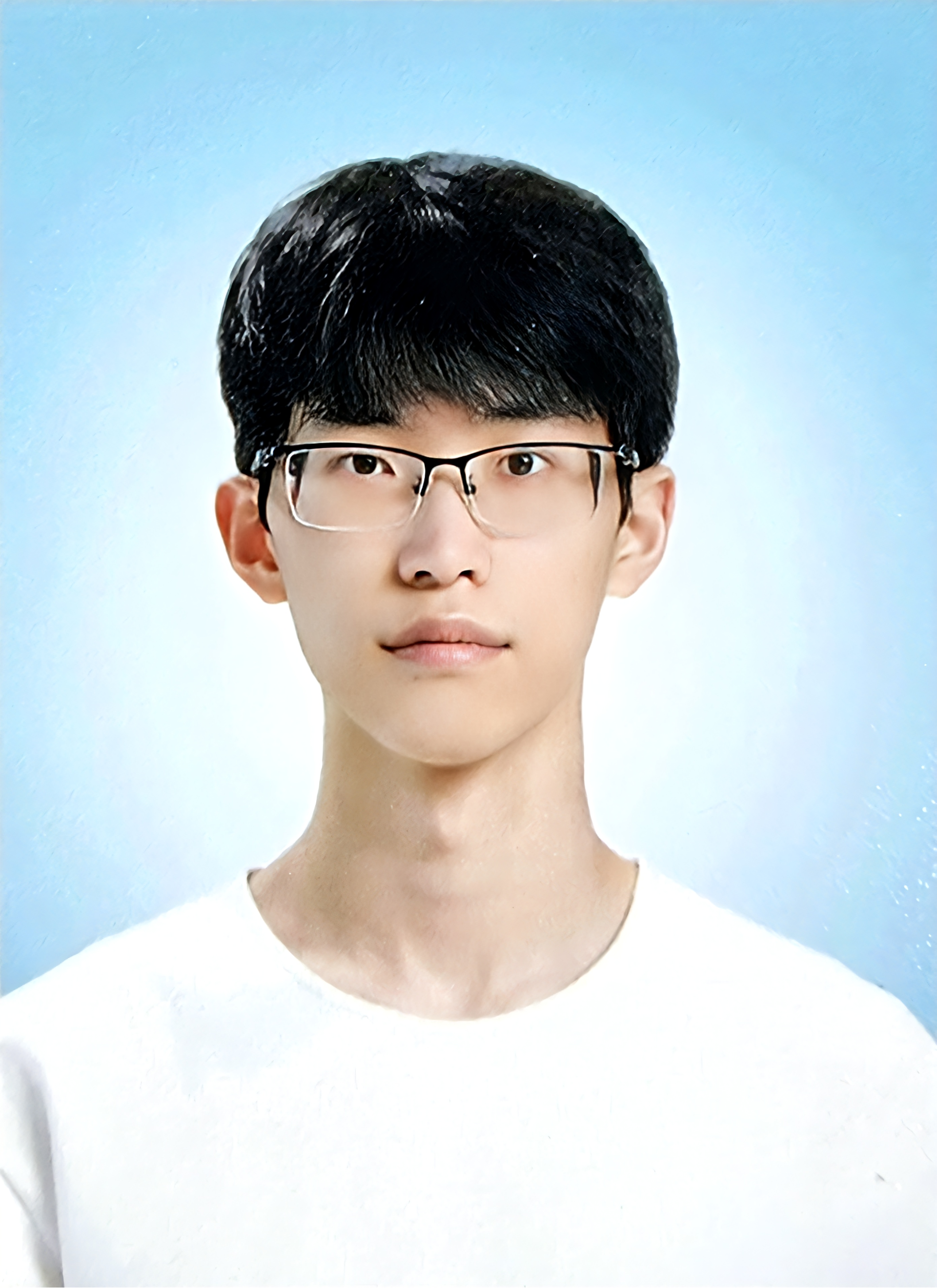}}]
{Hankyoul Ko} is currently an undergraduate student at the Department of Mechanical Engineering, Ulsan National Institute of Science and Technology, Ulsan, Republic of Korea. His research focuses on unmanned aerial vehicles, sensor fusion for localization, and robotic perception.
\end{IEEEbiography}

\begin{IEEEbiography}[{\includegraphics
[width=1in,height=1.25in,clip,
keepaspectratio]{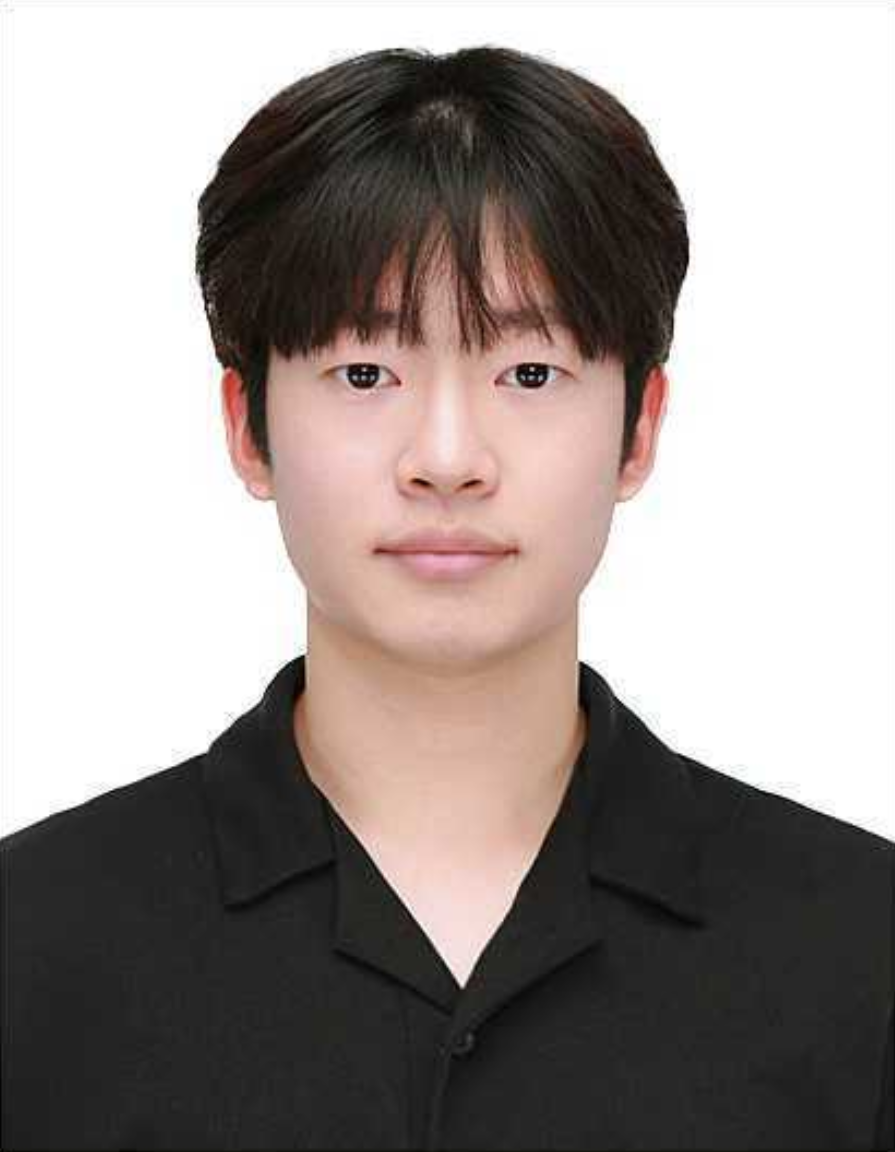}}]
{Hanse Cho} is currently a M.S. student at the Department of Mechanical Engineering, Ulsan National Institute of Science and Technology, Ulsan, Republic of Korea. He received a B.S. degree in Mechanical and Control Engineering from Handong Global University, Pohang, Republic of Korea, in 2024. His research focuses on flocking and swarm control.
\end{IEEEbiography}

\begin{IEEEbiography}[{\includegraphics
[width=1in,height=1.25in,clip,
keepaspectratio]{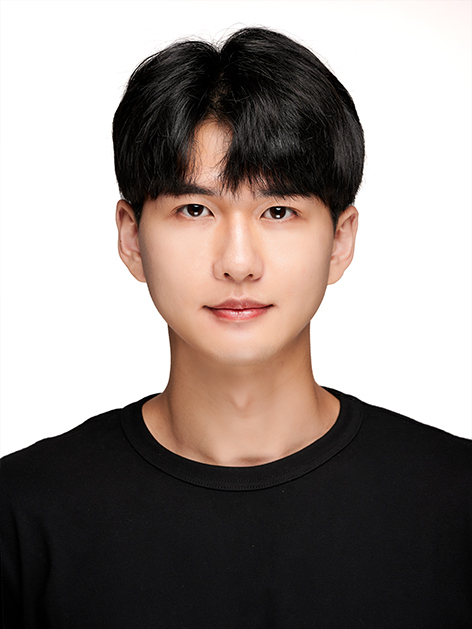}}]
{Changseung Kim} is currently a Ph.D. candidate at the Department of Mechanical Engineering, Ulsan National Institute of Science and Technology, Ulsan, Republic of Korea. He received a B.S. degree in Mechanical Engineering from Inha University, Incheon, Republic of Korea, in 2021. His research focuses on multi-sensor fusion simultaneous localization and mapping.
\end{IEEEbiography}

\begin{IEEEbiography}[{\includegraphics
[width=1in,height=1.25in,clip,
keepaspectratio]{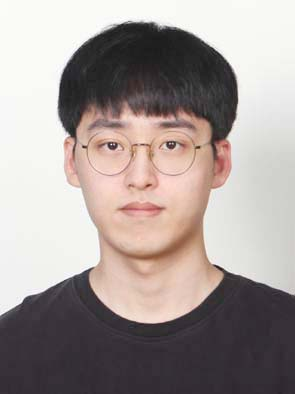}}]
{Seunghwan Kim} is currently a Ph.D. candidate at the Department of Mechanical Engineering, Ulsan National Institute of Science and Technology, Ulsan, Republic of Korea. He received a B.S. degree in Mechanical Engineering from Inha University, Incheon, Republic of Korea, in 2021. His research focuses on environmental monitoring, active perception, and autonomous exploration.
\end{IEEEbiography}

\begin{IEEEbiography}[{\includegraphics
[width=1in,height=1.25in,clip,
keepaspectratio]{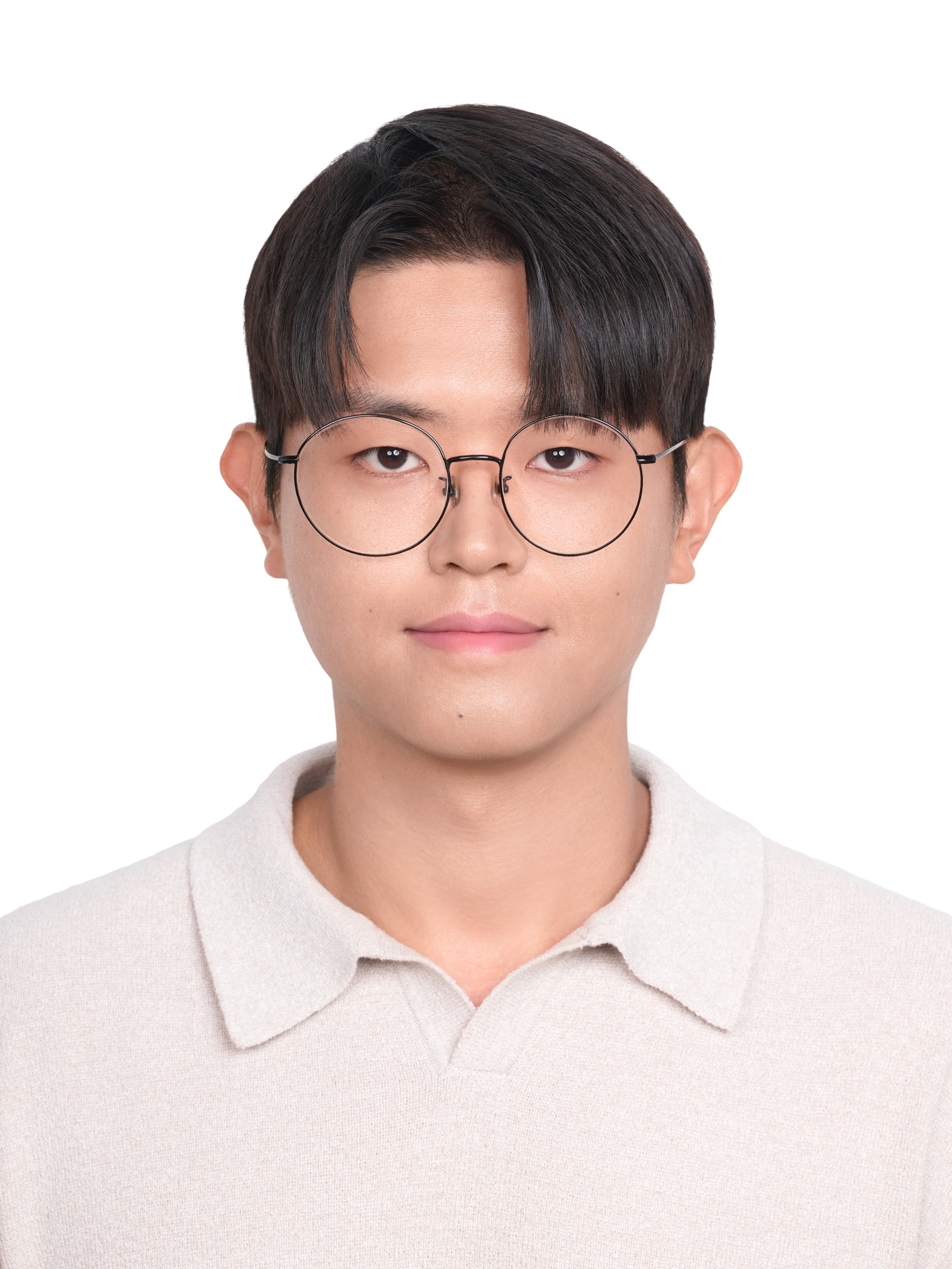}}]
{Jaemin Seo} is currently a Ph.D. candidate at the Department of Mechanical Engineering, Ulsan National Institute of Science and Technology, Ulsan, Republic of Korea. He received a B.S. degree in Aerospace Engineering from  Pusan National University, Busan, Republic of Korea, in 2019. His research focuses on environmental monitoring, informative path planning, and cooperative search strategy.
\end{IEEEbiography}

\begin{IEEEbiography}[{\includegraphics
[width=1in,height=1.25in,clip,
keepaspectratio]{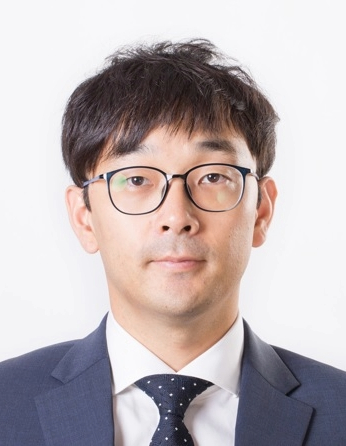}}]
{Hyondong Oh} is currently an Associate Professor at the Department of Mechanical Engineering, Korea Advanced Institute of Science and Technology, Daejeon, Republic of Korea. He received his B.S. and M.S. degrees in Aerospace Engineering from Korea Advanced Institute of Science and Technology, Daejeon, Republic of Korea, in 2004 and 2010, respectively, and a Ph.D. in Aerospace Engineering from Cranfield University, Cranfield, United Kingdom, in 2013. He was a Postdoctoral Researcher at the University of Surrey, Guildford, United Kingdom, from 2013 to 2014, an Assistant Professor at Loughborough University, Loughborough, United Kingdom, from 2014 to 2016, and an Associate Professor at the Ulsan National Institute of Science and Technology, Ulsan, Republic of Korea, from 2016 to 2025. His research interests include decision making for unmanned vehicles, cooperative control, path planning, nonlinear guidance and control, and sensor/information fusion.
\end{IEEEbiography}

\end{document}